\documentclass[10pt,journal,compsoc]{IEEEtran}

\ifCLASSOPTIONcompsoc
  \usepackage[nocompress]{cite}
\else
  \usepackage{cite}
\fi

\ifCLASSINFOpdf
  
\else
  
\fi

\usepackage{hyperref}
\hypersetup{
	colorlinks=true,
	citecolor=blue,
	linkcolor=blue,
	urlcolor=blue
}

\usepackage{amsmath}
\usepackage{amssymb}
\usepackage{pifont}
\usepackage{xcolor}
\definecolor{lightblue}{RGB}{70, 130, 230}
\definecolor{lightgreen}{RGB}{100, 160, 90}
\definecolor{lightpurple}{RGB}{148, 0, 211}
\definecolor{mygreen}{RGB}{68,137,51}

\usepackage[percent]{overpic}
\usepackage{multirow}
\usepackage{adjustbox}
\usepackage{soul}
\usepackage{arydshln}
\usepackage{graphicx}
\usepackage{threeparttable}
\usepackage{ragged2e}
\usepackage{colortbl}
\usepackage{caption}
\usepackage[table]{xcolor}
\usepackage{balance}

\definecolor{tablegreen}{RGB}{244, 199, 195} 
\definecolor{tableblue}{RGB}{218, 232, 252}
\definecolor{tablegray}{RGB}{213, 232, 212}
\definecolor{lightblue}{RGB}{235,243,252}
\definecolor{lightgreen}{RGB}{230, 255, 230}
\definecolor{lightgray}{RGB}{242, 242, 242}

\newcommand{\lb}{\cellcolor{lightblue}}
\newcommand{\lgr}{\cellcolor{lightgreen}}
\newcommand{\lga}{\cellcolor{lightgray}}

\begin{document}

\title{MoGen: A Unified Collaborative Framework for Controllable Multi-Object Image Generation}

\author{Yanfeng Li,~\IEEEmembership{}
        Yue Sun,~\IEEEmembership{}
        Keren Fu,~\IEEEmembership{}
        Sio-Kei Im,~\IEEEmembership{Senior Member,~IEEE},
        Xiaoming Liu,~\IEEEmembership{Fellow,~IEEE},
        Guangtao Zhai,~\IEEEmembership{Fellow,~IEEE},
        Xiaohong Liu,~\IEEEmembership{Member,~IEEE},
        and Tao Tan,~\IEEEmembership{Senior Member,~IEEE}
\IEEEcompsocitemizethanks{\IEEEcompsocthanksitem Yanfeng Li, Yue Sun, Sio-Kei Im and Tao Tan are with the Faculty of Applied Sciences, Macao Polytechnic University, Macao SAR, 999078, China (email: taotan@mpu.edu.mo).
\IEEEcompsocthanksitem Keren Fu is with the College of Computer Science, Sichuan University, Chengdu, 610065, China. (email: fkrsuper@scu.edu.cn)
\IEEEcompsocthanksitem Xiaoming Liu is with the Department of Computer Science and Engineering, Michigan State University, East Lansing, MI 48824, USA (email: liuxm@msu.edu)
\IEEEcompsocthanksitem Guangtao Zhai is with the School of Information Science and Electronic Engineering, Shanghai Jiao Tong University, Shanghai, 200240, China (email: zhaiguangtao@sjtu.edu.cn).
\IEEEcompsocthanksitem Xiaohong Liu is with the School of Computer Science, Shanghai Jiao Tong University, Shanghai, 200240, China (email: xiaohongliu@sjtu.edu.cn)
\IEEEcompsocthanksitem Corresponding Author: Xiaohong Liu, Tao Tan.}
}

% The paper headers
%\markboth{Journal of \LaTeX\ Class Files,~Vol.~14, No.~8, August~2015}%
%{Shell \MakeLowercase{\textit{et al.}}: Bare Advanced Demo of IEEEtran.cls for IEEE Computer Society Journals}

\IEEEtitleabstractindextext{%
\begin{abstract}
\justifying
Existing multi-object image generation methods face difficulties in achieving precise alignment between localized image generation regions and their corresponding semantics based on language descriptions, frequently resulting in inconsistent object quantities and attribute aliasing. To mitigate this limitation, mainstream approaches typically rely on external control signals to explicitly constrain the spatial layout, local semantic and visual attributes of images. However, this strong dependency makes the input format rigid, rendering it incompatible with the heterogeneous resource conditions of users and diverse constraint requirements. To address these challenges, we propose MoGen, a user-friendly multi-object image generation method. First, we design a Regional Semantic Anchor (RSA) module that precisely anchors phrase units in language descriptions to their corresponding image regions during the generation process, enabling text-to-image generation that follows quantity specifications for multiple objects. Building upon this foundation, we further introduce an Adaptive Multi-modal Guidance (AMG) module, which adaptively parses and integrates various combinations of multi-source control signals to formulate corresponding structured intent. This intent subsequently guides selective constraints on scene layouts and object attributes, achieving dynamic fine-grained control. Experimental results demonstrate that MoGen significantly outperforms existing methods in generation quality, quantity consistency, and fine-grained control, while exhibiting superior accessibility and control flexibility.
Code is available at: \url{https://github.com/Tear-kitty/MoGen/tree/master}.
\end{abstract}

\begin{IEEEkeywords}
Multi-object image generation, quantity consistency, fine-grained control, diffusion models.
\end{IEEEkeywords}}

\maketitle

\IEEEdisplaynontitleabstractindextext

\IEEEpeerreviewmaketitle

\ifCLASSOPTIONcompsoc
\IEEEraisesectionheading{\section{Introduction}\label{sec:introduction}}
\else
\section{Introduction}
\label{sec:introduction}
\fi
\IEEEPARstart{I}{mage} generation has emerged as one of the most transformative research areas in computer vision, attracting widespread attention from both academia and industry \cite{xiao2025omnigen, jayasumana2024rethinking, li2025comprehensive}. With significant progress in generative models, researchers can synthesize images with unprecedented quality and diversity. This technological breakthrough not only provides powerful creative tools for professional users but also enables non-professionals to generate high-quality, diverse visual content, substantially lowering the technical barriers to digital art creation \cite{xu2024ufogen, pang2025blenet, pang2024bio, cao2024partition, zhao2024mobilediffusion}.
\begin{figure}[!t]
\vspace{-2\baselineskip}
	\centering
	\begin{overpic}[width=0.50\textwidth, trim=15 0 15 15, clip]{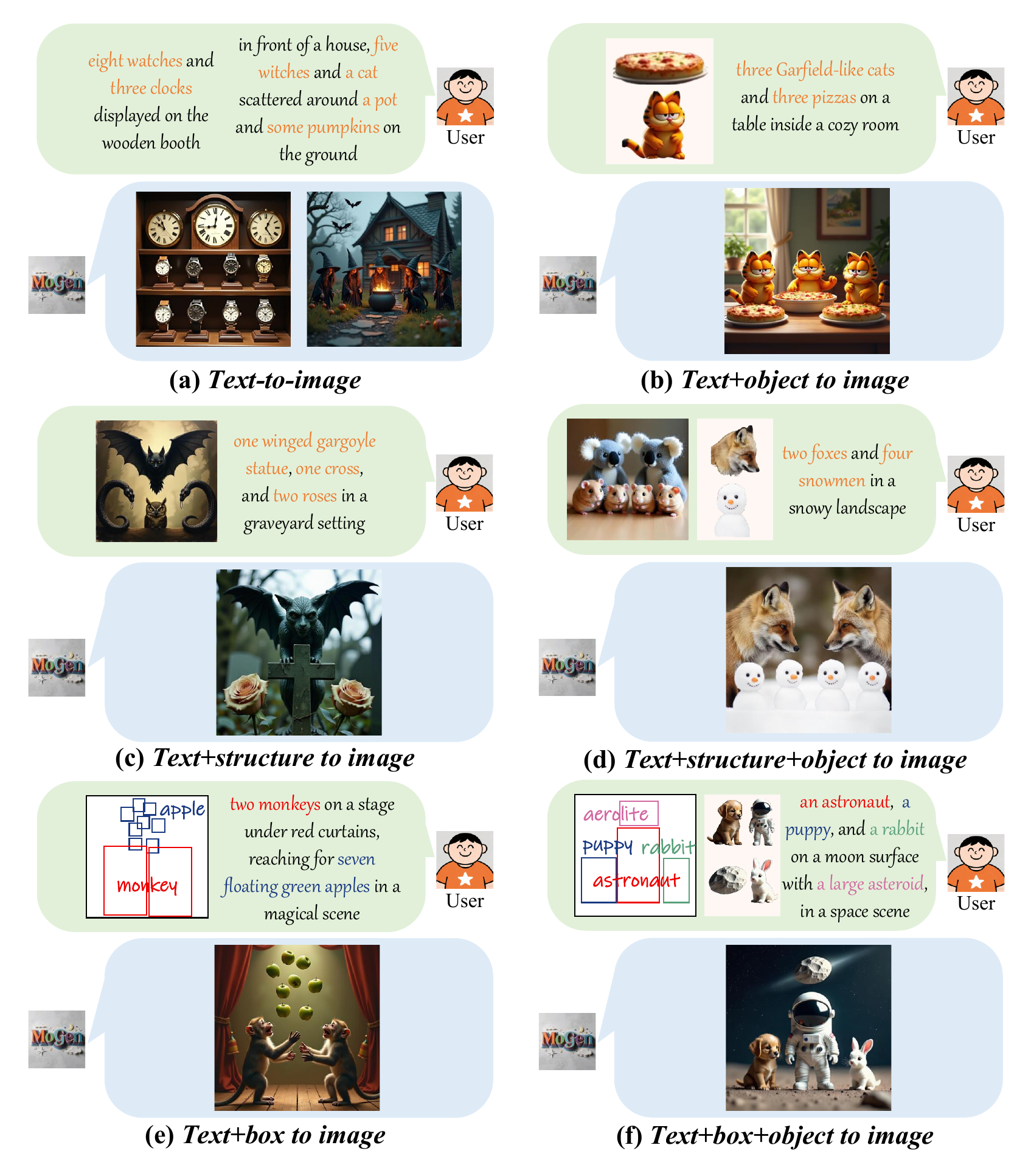}
	\end{overpic}
	\vspace{-19.5pt}
	\caption{\textbf{High-quality samples from our MoGen}. MoGen enables multi-object image generation with fine-grained control via text, optionally augmented with bounding boxes, structure references, and object references.}
	\label{fig:home_display}
\end{figure}
\begin{figure*}[!t]
	\centering
	\begin{overpic}[width=1.0\textwidth, trim=18 0 10 0, clip]{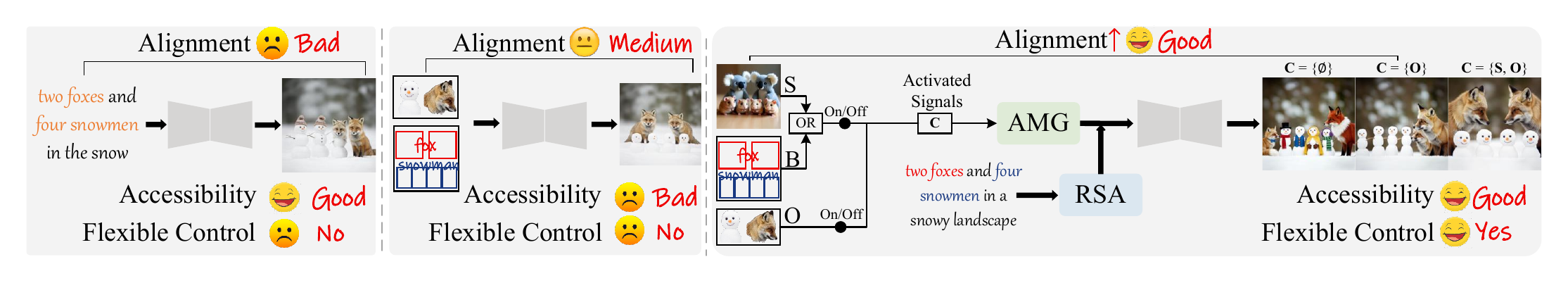}
		\put(0,-0.5){\footnotesize\parbox{4cm}{(a) Text-to-image method}}
		\put(23,-0.5){\footnotesize\parbox{4.5cm}{(b) Signal-based method}}
		\put(44.5,-0.5){\footnotesize\parbox{10cm}{(c) Proposed method MoGen [S: structure, O: object, B: box]}}
	\end{overpic}
	\vspace{-10pt}
	\caption{\textbf{The comparison between previous methods and our MoGen}. (a) Text-to-image methods offer high accessibility but often fail to ensure quantity consistency and lack fine-grained control over scene layouts and object attributes. (b) Signal-based methods enable fine-grained control, thereby improving quantity consistency, yet their reliance on the rigid input format limits accessibility and control flexibility. (c) MoGen enables text-to-image generation that follows quantity requirements, and can further integrate various combinations of multi-source control signals. Depending on the activated signals, the model imposes corresponding fine-grained control over scene layouts and object attributes. }
	\label{fig:defect}
	\vspace{-5pt}
\end{figure*}

Early research was pioneered by Stable Diffusion \cite{rombach2022high}, which was the first to perform the high-resolution image generation in the latent space. As the technology evolved, subsequent studies progressively expanded the boundaries of its applications \cite{flux2024, esser2024scaling, podell2023sdxl}. For instance, T-person \cite{liu2025t} and OmniGen \cite{xiao2025omnigen} achieved identity consistency in character generation, while CosiStory \cite{tewel2024training} further extended this paradigm to the generation of narratively coherent image sequences. However, existing methods still face critical challenges in multi-object composite scene generation, including difficulties in accurately constructing spatial layouts, inter-object relationships, and attribute constraints among multiple objects. These limitations significantly hinder scalability in the number of generable objects, thereby constraining the ability to meet diverse creative demands. 

In recent years, multi-object image generation has primarily explored text-to-image (T2I) paradigms due to their accessibility and ease of use \cite{zhong2023adapter}. Some works (such as Genartist \cite{wang2024genartist} and MuLan \cite{li2024mulan}) leverage large language models (LLMs) to enhance global text semantic modeling, improving quantity consistency in single-category or structurally simple scenarios. However, when descriptions contain multi-category objects with multi-attribute constraints, the global semantics provided by LLMs cannot be effectively grounded into phrase-level text semantics that capture individual object names and their modifiers. As a result, localized image generation regions often fail to precisely align with their corresponding text semantics, leading to semantic ambiguity and attribute aliasing, which in turn result in misgeneration. Additionally, textual descriptions lack explicit spatial semantics and visual references, leading to ambiguous representations of key visual elements such as appearance prototypes and relative positions, which inadequately support layout control and attribute restoration.

To address these problems, another research introduces human-predefined signals as external controls to explicitly constrain spatial layouts and visual attributes of localized regions. For example, LLM-grounded \cite{fan2024prompt} and BoxDiff \cite{xie2023boxdiff} use bounding boxes to drive generation for precise layouts, significantly improving quantity consistency. Building on this, Emu2 \cite{sun2024generative} and CompAgent \cite{wang2024divide} further incorporate object references to achieve appearance consistency and attribute controllability. Despite their effectiveness, these methods face two key limitations: first, the high cost of constructing control signals leads to method designs that favor single control paradigms; second, they lack the capability to adaptively parse and integrate structured intent from various combinations of multi-source control signals, resulting in rigid input formats. These factors weaken accessibility and control flexibility, thereby limiting adaptability to heterogeneous resource conditions and diverse control requirements, ultimately compromising user convenience.

We propose MoGen, as shown in Fig.~\ref{fig:home_display}, a user-friendly multi-object image generation method. First, MoGen introduces a Regional Semantic Anchor (RSA) module to alleviate the misalignment between localized image generation regions and their corresponding text semantics in multiple categories scenarios with coexisting attributes. This enables the independence of semantics and attributes for each image region, thereby ensuring the text-to-image generation that follows quantity requirements and providing an accessible and extensible generation method for users.
Building upon this, MoGen further incorporates the Adaptive Multi-modal Guidance (AMG) module to achieve adaptive parsing and integration of various combinations of multi-source control signals, thereby allowing flexible incorporation of bounding boxes, structure references, and object references. These signals can be utilized individually or in combinations, and according to the activated signals, AMG formulates their corresponding structured intent to dynamically impose fine-grained control over scene layouts and object attributes.
This design enables MoGen to enhance generation controllability and control flexibility while maintaining accessibility, effectively adapting to varying user requirements and resource conditions. Fig.~\ref{fig:defect} demonstrates the comparison between previous methods and MoGen.
The specific implementation involves three key components:

\emph{1)} The RSA module, incorporates two key designs: \emph{(i)} a task-oriented Semantic Parser is introduced to supersede reliance on LLMs by modeling global text semantics from text embeddings output by text encoder and further decomposing these embeddings into phrase-level text semantics. \emph{(ii)} a synergistic utilization mechanism is proposed, where global text semantics steer all U-Net blocks for structural stability in image generation, while phrase-level text semantics are integrated into the layout block \cite{wang2024instantstyle} to further establish precise alignment between localized generation regions and their corresponding text semantics. These designs effectively ensure quantity consistency.
\begin{table*}[!t]
	\centering
	\caption{\textbf{An overview of previous studies.} 
		{Quantity} and {Ada Struc.} indicate whether quantity consistency and adaptive structured intent parsing mechanisms are incorporated, respectively. 
		{w/o LLM} indicates whether the method operates without relying on LLMs. Structure reference and bounding box are not necessarily used simultaneously.}
	\vspace{-5pt}
	\renewcommand{\arraystretch}{1.2}
	\setlength{\tabcolsep}{6pt}
	\resizebox{\textwidth}{!}{
		\begin{threeparttable}
			\begin{tabular}{lccccccccccc}
				\hline
				\rowcolor{white}
				\textbf{Method} & & \textbf{Quantity} & \textbf{Ada Struc.} & \textbf{w/o LLM} & \textbf{Text} & \textbf{Text+structure\ } & \textbf{Text+object} & \textbf{Text+box} & \textbf{Text+object+box} & \textbf{Text+object+structure\ } \\ 
				\hline
				\rowcolor{lightblue} SDXL & \cite{podell2023sdxl} &  &  & $\checkmark$ & $\checkmark$ &  &  &  &  &  \\
				FLUX & \cite{flux2024} &  &  & $\checkmark$ & $\checkmark$ &  &  &  &  &  \\
				\rowcolor{lightblue} Emu2 & \cite{sun2024generative} &  &  &  & $\checkmark$ & $\checkmark$ &  & $\checkmark$ & $\checkmark$ &  \\
				Omigen2 & \cite{wu2025omnigen2} &  &  &  & $\checkmark$ & $\checkmark$ & $\checkmark$ &  &  &  \\
				\rowcolor{lightblue} Bounded-attention & \cite{dahary2024yourself} &  &  & $\checkmark$ &  &  &  & $\checkmark$ &  &  \\
				Xverse & \cite{chen2025xverse} &  &  & $\checkmark$ &  &  & $\checkmark$ &  &  &  \\
				\rowcolor{lightblue} MS-diffusion & \cite{wangms} &  &  & $\checkmark$ &  &  & $\checkmark$ &  & $\checkmark$ &  \\
				StableFlow & \cite{avrahami2025stable} &  &  & $\checkmark$ &  & $\checkmark$ &  &  &  &  \\
				\rowcolor{lightblue} Flow-Edit & \cite{kulikov2024flowedit} &  &  & $\checkmark$ &  & $\checkmark$ &  &  &  &  \\
				MoEdit & \cite{li2025moedit} & $\checkmark$ &  & $\checkmark$ &  & $\checkmark$ &  &  &  &  \\
				\rowcolor{lightblue} \textbf{MoGen (Ours)} &  & $\checkmark$ & $\checkmark$ & $\checkmark$ & $\checkmark$ & $\checkmark$ & $\checkmark$ & $\checkmark$ & $\checkmark$ & $\checkmark$ \\ 
				\hline
			\end{tabular}
		\end{threeparttable}
	}
	\label{tab:previous_method}
\end{table*}

\emph{2)} The AMG module, comprises two hierarchical components: \emph{(i)} Signal Encoder, which adaptively parses activated heterogeneous control signals and projects them into a unified feature representation. \emph{(ii)} Adaptive Controller, which integrates inter-signal correlations and dependencies based on this representation, thereby formulating corresponding structured intent which modulates the spatial scope and attribute emphasis of constraints, ultimately enabling flexible and controllable fine-grained image generation.

\emph{3)} Multi-object Control Annotation (MoCA) benchmark, provides a data foundation for MoGen. Through its open-source release, the MoCA aims to reduce data construction costs for subsequent research. MoCA consists of multi-object images with cross-modal annotations, including bounding boxes, descriptive text, structure references, and object references. 
%The data is organized as follows: each image contains a descriptive text and structural reference, with $1$--$15$ annotated objects. Each object includes a corresponding bounding box and object reference. 
This design provides explicit spatial priors for layout control and visual prototypes for attribute constraints.

This work is a substantial extension of On Learning Quantity Perception for Multi-object Image Editing (MoEdit) \cite{li2025moedit} published in the Proceedings of the IEEE/CVF Conference on Computer Vision and Pattern Recognition (CVPR 2025). Compared with the conference version, this paper introduces three major improvements:
(1) Paradigm extension from editing to generation. Most importantly, we significantly enhance structural stability in generation process and achieve precise alignment between localized image regions and their corresponding phrase-level text semantics. This extends our original multi-object editing framework \cite{li2025moedit}, which was constrained by visual anchors, into a text-to-multi-object image generation framework with greater expressive freedom.
(2) Flexible fine-grained control. 
We introduce an adaptive controllable network that allows to activate various combinations of multi-source control signals, achieving dynamic fine-grained control. Unlike MoEdit \cite{li2025moedit}, this work removes the rigid dependency on reference images, allowing adaptation to varying resource conditions and constrain requirements.
(3) Data and annotation system reconstruction. We scale up the original editing dataset and introduce new annotation dimensions including bounding boxes, object references, and structure references for image generation task, providing a solid benchmark for fine-grained layout and attribute control.

In summary, our contributions are as follows:
\begin{itemize}
	\item We propose MoGen, a user-friendly multi-object generation method that balances controllability, accessibility, and flexibility, while effectively adapting to diverse user requirements and resource conditions.
	\item We introduce the RSA module, which generates images that follow quantity requirements specified by the input text, enabling accessible generation. We also propose the AMG module, which adapts to various combinations of multi-source control signals to achieve dynamic fine-grained control.
	\item To establish a data foundation for fine-grained control and alleviate the constraint of data construction costs on subsequent research, we collect and open-source the MoCA benchmark, providing images, text pairs, and corresponding control signal annotations.
	\item Compared to existing methods, MoGen achieves superior performance in multi-object generation quality, quantity consistency, and fine-grained control, while providing more diverse control signals and input configurations, improving user convenience.
\end{itemize}
\begin{figure*}[!t]
	\centering
	\begin{overpic}[width=1.0\textwidth, trim=18 15 15 0, clip]{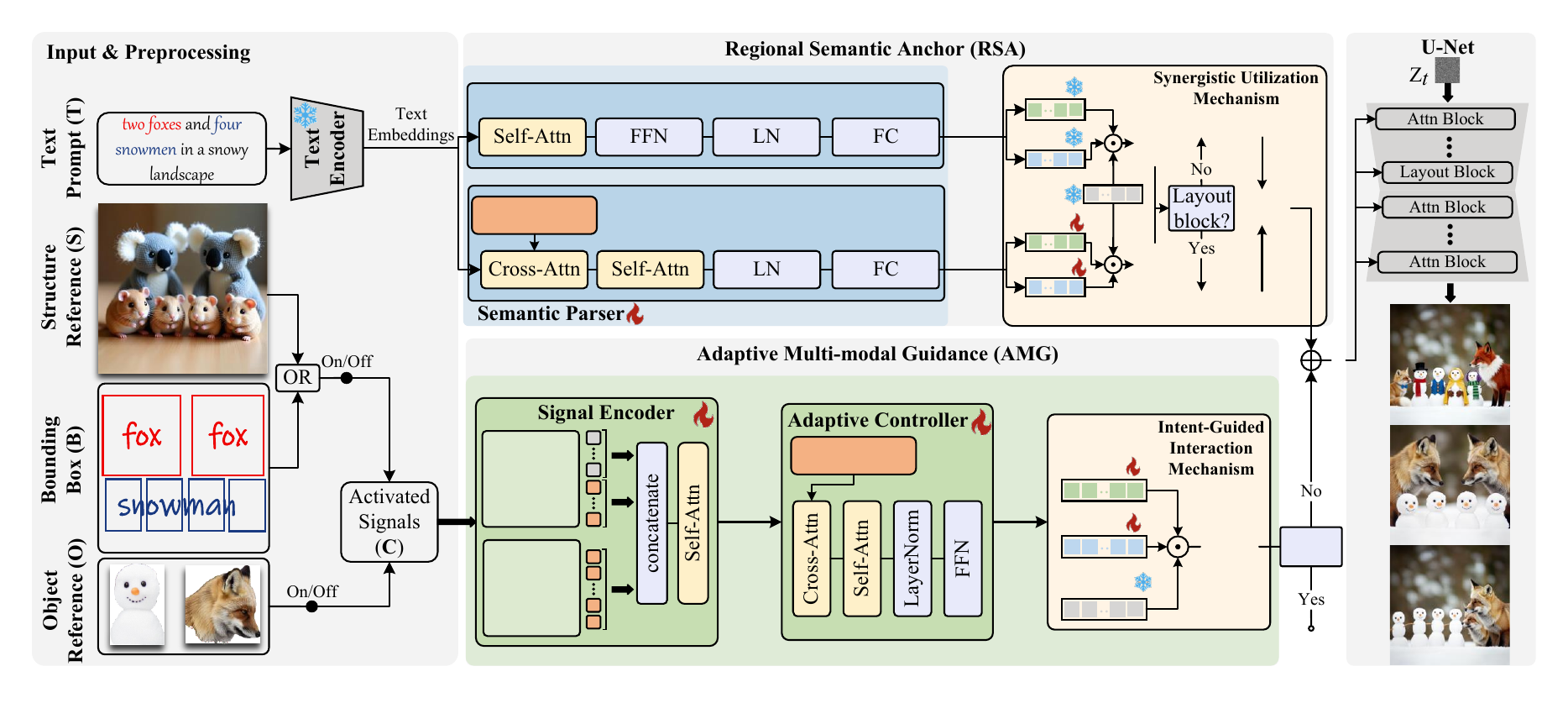}
        \put(39.3,37.7){\scriptsize\parbox{4cm}{\textbf{Global Branch} ($\mathbf{SP}_{glob}$)}}
        \put(39.3,30.1){\scriptsize\parbox{4cm}{\textbf{Phrase-level Branch} ($\mathbf{SP}_{phr}$)}}
        \put(23.5,33.7){\scriptsize\parbox{4cm}{$\mathbf{T}_{emb}$}}
		\put(29.5,30.1){\scriptsize\parbox{4cm}{Query $\mathbf{Q}_{phr}$}}
        \put(22.0,12.8){\scriptsize\parbox{4cm}{\centering Image\\ Encoder\\ ($\mathbf{E}_{img}$)}}
        \put(22.0,5.5){\scriptsize\parbox{4cm}{\centering Box\\ Encoder\\ ($\mathbf{E}_{box}$)}}
        \put(50.8,14.3){\scriptsize\parbox{4cm}{Query $\mathbf{Q}_{str}$}}
        \put(45.6,11){\tiny\parbox{4cm}{$\mathbf{C}_{unif}$}}
        \put(64.1,11){\tiny\parbox{4cm}{$\mathbf{C}_{str}$}}
        \put(60.6,36.2){\scriptsize\parbox{4cm}{$\mathbf{T}_{glob}$}}
        \put(60.6,27.4){\scriptsize\parbox{4cm}{$\mathbf{T}_{phr}$}}
        \put(64.7,29.9){\scriptsize\parbox{4cm}{$\mathbf{K}_{phr}$}}
        \put(64.7,26.9){\scriptsize\parbox{4cm}{$\mathbf{V}_{phr}$}}
        \put(64.7,31.8){\scriptsize\parbox{4cm}{$\mathbf{Q}_{net}$}}
        \put(64.7,38.6){\scriptsize\parbox{4cm}{$\mathbf{K}_{glob}$}}
        \put(64.7,35.4){\scriptsize\parbox{4cm}{$\mathbf{V}_{glob}$}}
        \put(75.5,36.4){\tiny\parbox{4cm}{($\mathbf{\hat{V}}_{rsa}$=$\mathbf{V}_{glob}^{'}$)}}
        \put(73,34.8){\tiny\parbox{4cm}{$\mathbf{V}_{glob}^{'}$}}
        \put(72.6,27.6){\tiny\parbox{4cm}{$\mathbf{V}_{phr}^{'}$}}
        \put(80.3,30.7){\tiny\parbox{4cm}{$\mathbf{\hat{V}}_{rsa}$}}
        \put(72.5,24.5){\tiny\parbox{4cm}{($\mathbf{\hat{V}}_{rsa}$=$\mathbf{V}_{glob}^{'}$+$\mathbf{V}_{phr}^{'}$)}}
        \put(85.3,18.1){\rotatebox{90}{\scriptsize\parbox{4cm}{$\mathbf{\hat{V}}_{mogen}$}}}
        \put(69.0,13.6){\scriptsize\parbox{4cm}{$\mathbf{K}_{str}$}}
        \put(69.0,9.8){\scriptsize\parbox{4cm}{$\mathbf{V}_{str}$}}
        \put(69.0,5.8){\scriptsize\parbox{4cm}{$\mathbf{Q}_{net}$}}
        \put(77.3,8.3){\scriptsize\parbox{4cm}{$\mathbf{\hat{V}}_{amg}$}}
        \put(88,7.4){\rotatebox{90}{\scriptsize\parbox{4cm}{Generated Image ($\mathbf{I}_{g}$)}}}
        \put(98.2,2.8){\rotatebox{90}{\tiny\parbox{4cm}{$\mathbf{C}$=$\{\mathbf{O}\}$}}}
        \put(98.2,10.2){\rotatebox{90}{\tiny\parbox{4cm}{$\mathbf{C}$=$\{\mathbf{S},\mathbf{O}\}$}}}
        \put(98.2,18.5){\rotatebox{90}{\tiny\parbox{4cm}{$\mathbf{C}$=$\{\varnothing\}$}}}
        \put(29.1,15.6){\tiny\parbox{4cm}{\centering $\mathbf{F}_s(\varnothing)$}}
        \put(38.4,12.2){\tiny\parbox{4cm}{$\mathbf{F}_o$}}
        \put(38.4,6.7){\tiny\parbox{4cm}{$\mathbf{F}_b$}}
        \put(73.8,8.3){\tiny\parbox{4cm}{\centering$\mathbf{C}$ = \\$\{\varnothing\}$?}}
	\end{overpic}
	\vspace{-15pt}
	\caption{\textbf{The framework of MoGen}. 
    Regional Semantic Anchor (RSA) module decomposes input text prompts into global text semantics $\mathbf{T}_{glob}$ and phrase-level text semantics $\mathbf{T}_{phr}$. Then, Synergistic Utilization Mechanism broadcasts $\mathbf{V}_{glob}'$ (derived from $\mathbf{T}_{glob}$) across all U-Net blocks to ensure structural stability of generated images, while integrating $\mathbf{V}_{phr}'$ (derived from $\mathbf{T}_{phr}$) to layout blocks \cite{li2025moedit} to further enforce alignment between local image regions and their corresponding text semantics, thereby ensuring quantity consistency in text-to-multi-object image generation. For fine-grained control, Adaptive Multi-modal Guidance (AMG) module is employed. Signal Encoder encodes activated signals $\mathbf{C}$ into features ${\mathbf{F}_{i}}$, $i\in\{s,o,b\}$, where ${(\varnothing)}$ represents null embeddings, and further forms a unified feature representation $\mathbf{C}_{unif}$. An Adaptive Controller then generates a structured intent $\mathbf{C}_{str}$, which comprises spatial layout and visual attribute specifications of constraints, and is propagated into U-Net via Intent-Guided Interaction Mechanism. $\mathbf{Q}_{net}$ denotes pre-trained queries in U-Net. 
	}
	\label{fig:framework}
	\vspace{-0pt}
\end{figure*}
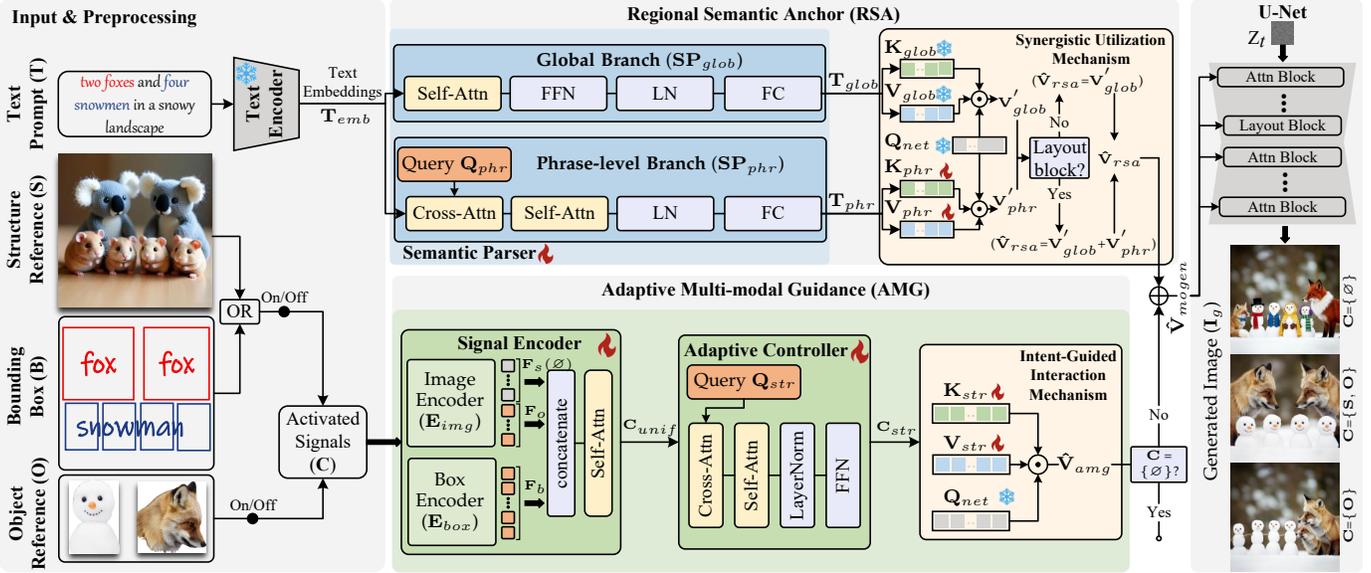
\section{Related Work}
\subsection{Text-to-image Diffusion Models}
Diffusion models have become the dominant paradigm for T2I generation, with representative models like SDXL \cite{podell2023sdxl}, FLUX \cite{flux2024}, and HunyuanDiT \cite{li2024hunyuan} serving as widely adopted baselines for generating high-quality, semantically consistent images. These mature models have been extended to diverse applications, including image editing \cite{huang2025diffusion, sheynin2024emu, feng2025dit4edit}, inpainting \cite{wasserman2025paint, wang2025towards, xie2025turbofill}, and creative synthesis \cite{xie2025sana, kim2025diffusehigh, agarwal2025image}. However, a critical limitation persists: they struggle to accurately follow quantity specifications in text prompts \cite{binyamin2025make}. This challenge becomes more pronounced for larger quantities, revealing a fundamental gap in current T2I capabilities.
\subsection{Multi-object Image Generation}
LLMs are widely adopted for multi-object generation due to their text semantic parsing capability, improving quantity consistency \cite{feng2023layoutgpt, cheng2024theatergen}. Direct approaches like LLM4GEN \cite{liu2025llm4gen}, DiffusionGPT \cite{qin2024diffusiongpt}, and Emu2 \cite{sun2024generative} leverage LLMs for global text semantic modeling. However, they struggle to extract phrase-level text semantics, leading to imprecise alignment between text semantics and target regions, with generation scenarios confined to single-category or structurally simple settings.
Methods like RPG \cite{yang2024mastering} and LLM-grounded Diffusion \cite{lian2023llm} attempt to address these limitations by employing LLMs as global planners for region partitioning and task decomposition, combined with region-specific diffusion models. However, these approaches still suffer from dilution of quantity information during semantic decomposition, leading to persistent attribute alignment confusion and object quantity deviation. Furthermore, they lack explicit spatial semantics and visual reference mechanisms for precise fine-grained scene layout and object attribute expression \cite{zhang2025layercraft}.
\subsection{Controllable Multi-Object Image Generation}
To address the limitations of text-based generation schemes, StableFlow \cite{avrahami2025stable} and FlowEdit \cite{kulikov2024flowedit} employ overall image structure as global consistency constraints to achieve fine-grained layout control, but their expressive freedom is significantly restricted. In contrast, Bounded-attention \cite{dahary2024yourself} and HiCo \cite{ma2024hico} utilize bounding boxes to impose spatial constraints, improving expressive freedom while ensuring layout controllability. Another category of methods (such as Xverse \cite{chen2025xverse} and Omnigen2 \cite{wu2025omnigen2}) focuses on introducing object references, achieving attribute-level constraints. Subsequently, MultiGen \cite{wu2024multigen} and MS-diffusion \cite{wangms} introduce multi-scale conditional encoding, enabling joint modeling of spatial constraints and object references and thereby further improving controllability. Although these methods realize specific fine-grained control and thereby improve quantity consistency, they rely on predefined input formats, limiting user accessibility and control flexibility.
In this work, we propose MoGen, which allows text-driven generation that follows quantity requirements for enhanced accessibility, while enabling flexible integration of multi-source control signals. The key innovation lies in dynamic constraint adjustment, which adapts fine-grained control mechanisms based on activated control signals. This design balances controllability, flexibility and accessibility, providing a user-friendly multi-object image generation method. Previous methods are summarized in Table~\ref{tab:previous_method}.
\section{Method}
\subsection{Overall}
In this work, we propose MoGen, denoted as ${\mathcal{G}}$, and illustrated in Fig.~\ref{fig:framework}, a user-friendly multi-object image generation method that allows users to select appropriate generation strategies based on task requirements and available resources. Specifically, MoGen enables text-to-image generation that follows quantity requirements. When fine-grained control is required, the model allows flexible activation of multi-source control signals to dynamically constrain the generation process. The generation process is formalized as:
\begin{equation}
	\begin{aligned}
		\mathbf{I}_g = \mathcal{G}(\mathbf{T}, \mathbf{C}),
	\end{aligned}
	\label{eq:MoGen}
\end{equation}
where $\mathbf{I}_g$ denotes the generated image, $\mathbf{T}$ represents the input text prompt, and $\mathbf{C}$ represents the activated control signals, with $\mathbf{C}\subseteq\mathbf{X}$ and $\{\mathbf{S},\mathbf{B}\}\nsubseteq\mathbf{C}$. The set $\mathbf{X}=\{\mathbf{S},\mathbf{O},\mathbf{B}\}$ contains three types of control signals: structure reference ($\mathbf{S}$), object reference ($\mathbf{O}$), and bounding box ($\mathbf{B}$).

The method comprises two main modules: the Regional Semantic Anchor (RSA) module and the Adaptive Multi-modal Guidance (AMG) module. 
The RSA module models global text semantics $\mathbf{T}_{glob}$ based on text embeddings $\mathbf{T}_{emb}$ from the text encoder to guide the overall structure of generated images, while further decomposing $\mathbf{T}_{emb}$ into phrase-level text semantics $\mathbf{T}_{phr}$ for precise alignment with localized image regions. This design effectively prevents semantic ambiguity and attribute aliasing among image regions, ensuring reliable and quantity-consistent generation.
The AMG module adaptively parses and integrates various combinations of multi-source control signals $\mathbf{C}$ to extract corresponding structured intent $\mathbf{C}_{str}$. Based on this intent, the module modulates the spatial scope and attribute emphasis of constraints, applying dynamic fine-grained control over scene layouts and object attributes. 

%The subsequent sections are organized as follows: Section~\ref{sec:RSA} details the specific mechanisms of RSA module, Section~\ref{sec:Adaptive Multi-modal Guidance} introduces the working principles of AMG module.
%and Section~\ref{sec:MoCA} describes the construction process and content of the MoCA benchmark.
\subsection{Regional Semantic Anchor}
\label{sec:RSA}
The RSA module takes text embeddings $\mathbf{T}_{emb}\in\mathbb{R}^{L_{emb}\times d}$ derived from the text encoder as input. Unlike conventional approaches that directly utilize $\mathbf{T}_{emb}$ for guidance, we aim to extract multi-level text semantics from the information-rich yet redundant $\mathbf{T}_{emb}$. By effectively injecting these text semantics, we achieve multi-object image generation that precisely follows user-specified quantity requirements. Specifically, the module comprises two core components: {Semantic Parser} and {Synergistic Utilization Mechanism}. First, we introduce a Semantic Parser to model global text semantics $\mathbf{T}_{glob}\in\mathbb{R}^{L_{emb}\times d}$ and extract phrase-level text semantics $\mathbf{T}_{phr}\in\mathbb{R}^{L_{phr}\times d}$ from $\mathbf{T}_{emb}$. 
%Subsequently, diverging from standard paradigms that concatenate semantic representations for uniform injection into all U-Net blocks, we devise a Synergistic Utilization Mechanism characterized by a differentiated injection strategy.  
Subsequently, we devise a Synergistic Utilization Mechanism, in which $\mathbf{T}_{glob}$ guides the overall generation structure of the image, while $\mathbf{T}_{phr}$ is further aligned with corresponding localized image regions to mitigate the text semantic ambiguity and attribute aliasing observed during text-to-multi-object generation, ensuring quantity consistency.\\

\noindent\textbf{Semantic Parser.}
As illustrated in Fig.~\ref{fig:framework}, the Semantic Parser $\mathbf{SP}$ employs two parallel computational branches: Global Branch ($\mathbf{SP}_{glob}$) and Phrase-level Branch ($\mathbf{SP}_{phr}$). 
$\mathbf{SP}_{glob}$: addressing the issue where the $\mathbf{T}_{emb}$ extracted by text encoder (or computationally expensive LLMs) often contains significant redundancy or informational noise, $\mathbf{SP}_{glob}$ introduces a self-attention mechanism to filter noise and compress redundancy, thereby modeling compact, controllable global text semantics $\mathbf{T}_{glob}$ that is aligned with the generation task. This $\mathbf{T}_{glob}$ significantly enhances the visual fidelity and structural stability of generated images, while also improving quantity consistency. Formally:
\begin{equation}
	\begin{aligned}
		\mathbf{T}_{glob} = \mathbf{SP}_{glob}(\mathbf{T}_{emb}).
	\end{aligned}
\end{equation}
The specific computation of $\mathbf{SP}_{glob}$ is as follows:
\begin{equation}
	\begin{aligned}
        &\mathbf{T}_{emb}^{'} = \texttt{FFN}(\texttt{SelfAttn}(\mathbf{T}_{emb})),\\
        &\mathbf{T}_{glob} = \texttt{FC}(\texttt{LN}(\mathbf{T}_{emb}^{'})),\\
	\end{aligned}
    \label{eq:Global Semantic Branch}
\end{equation}
where, $\texttt{SelfAttn}$ refers to the self-attention operation, $\texttt{FFN}$ denotes feed-forward network, $\texttt{LN}$ denotes layer normalization, $\texttt{FC}$ denotes fully connected layer, $\mathbf{T}_{emb}^{'}\in\mathbb{R}^{77\times d}$ denotes the intermediate variable in $\mathbf{SP}_{glob}$.

$\mathbf{SP}_{phr}$: given that $\mathbf{T}_{emb}$ typically functions as a highly coupled holistic sentence vector, this branch utilizes a cross-attention mechanism with learnable queries $\mathbf{Q}_{phr}\in\mathbb{R}^{L_{phr}\times d}$ to decompose $\mathbf{T}_{emb}$ into phrase-level text semantics $\mathbf{T}_{phr}$. Compared to more redundant $\mathbf{T}_{emb}$, $\mathbf{T}_{phr}$, consisting of multiple phrase units with independent semantics, establishes a foundation for the precise alignment between localized generated regions and specific text semantics, particularly in scenarios characterized by high spatial-semantic heterogeneity. The computation is formalized as:
\begin{equation}
	\begin{aligned}
		\mathbf{T}_{phr} = \mathbf{SP}_{phr}(\mathbf{T}_{emb}).
	\end{aligned}
\end{equation}
The specific computation of $\mathbf{SP}_{phr}$ is as follows:
\begin{equation}
	\begin{aligned}
        &\mathbf{T}_{emb}^{''} = \texttt{SelfAttn}(\texttt{CrossAttn}(\mathbf{Q}_{phr}, \mathbf{K}_{emb}, \mathbf{V}_{emb})),\\
		&\mathbf{T}_{phr} = \texttt{FC}(\texttt{LN}(\mathbf{T}_{emb}^{''})), 
	\end{aligned}
	\label{eq:Phrase-level Attribute Branch}
\end{equation}
where, $\texttt{CrossAttn}$ refers to the cross-attention operation, $\mathbf{T}_{emb}^{''}\in\mathbb{R}^{L_{phr}\times d}$ denotes the intermediate variable in $\mathbf{SP}_{phr}$, $\mathbf{K}_{emb}$ and $\mathbf{V}_{emb}$ are derived from $\mathbf{T}_{emb}$.\\
% \begin{figure}[!t]
% 	\centering
% 	\begin{overpic}[width=0.5\textwidth, trim=18 10 0 25, clip]{fig/feature_injection.pdf}
% 		\put(25.5,8){\small\parbox{4cm}{$\mathbf{T}_{glob}$}}
% 		\put(55,7.5){\small\parbox{4cm}{$\mathbf{T}_{phr}$}}
% 		\put(84,8){\parbox{4cm}{$p_s$}}
% 		\put(5.5,8){\small\parbox{4cm}{$\mathbf{T}_{emb}$}}
% 	\end{overpic}
% 	\vspace{-15pt}
% 	\caption{\textbf{Illustration of injecting $\mathbf{T}_{glob}$, $\mathbf{T}_{phr}$ and $p_s$ into each layer of MoGen}. The global semantics $t_g$ directly replaces the original text embedding $t_e$ and achieves feature injection through existing cross-attention layers within the U-Net. In contrast, we introduce learnable cross-attention layers in the layout block \cite{wang2024instantstyle} to inject $t_p$. Building upon this design, we further incorporate additional learnable cross-attention layers to inject the structured intent $p_s$, thereby achieving precisely fine-grained control over scene layouts and object attributes. Notably, they all share the existing queries $Q$ from the U-Net, ensuring consistent feature alignment across different levels of semantic control.}
% 	\label{fig:feature_interaction}
% 	\vspace{-5pt}
% \end{figure}

\noindent\textbf{Synergistic Utilization Mechanism.}
To balance the effectiveness of quantity-consistent generation with pretrained fidelity, we propose a Synergistic Utilization Mechanism. The core of this mechanism lies in distinguishing the functional roles of $\mathbf{T}_{glob}$ and $\mathbf{T}_{phr}$. By designing distinct injection strategies for each, we avoid excessive disruption to pre-trained feature distributions common in traditional methods. Specifically, $\mathbf{T}_{glob}$ serves as a global baseline to provide stable structural guidance for image generation, whereas $\mathbf{T}_{phr}$ is introduced to enable the precise alignment between localized image regions and their corresponding text semantics. Therefore, as illustrated in Fig.~\ref{fig:framework}, $\mathbf{T}_{glob}$ performs semantic injection on all U-Net blocks via pre-trained cross-attention layers, which use queries $\mathbf{Q}_{net}\in\mathbb{R}^{L_{net}\times d_{net}}$:
\begin{equation}
	\begin{aligned}
		\mathbf{V}_{glob}^{'} = \texttt{CrossAttn}(\mathbf{Q}_{net}, \mathbf{K}_{glob}, \mathbf{V}_{glob}),
	\end{aligned}
	\label{eq:global Synergistic Utilization Mechanism}
\end{equation}
where, $\mathbf{K}_{glob}$ and $\mathbf{V}_{glob}$ are derived from $\mathbf{T}_{glob}$, $\mathbf{V}_{glob}^{'}\in\mathbb{R}^{L_{net}\times d_{net}}$. In contrast, $\mathbf{T}_{phr}$ is integrated via trainable cross-attention layers only appended to the layout block \cite{li2025moedit} (fourth U-Net block) to provide guidance:
\begin{equation}
	\begin{aligned}
		\mathbf{V}_{phr}^{'} = \texttt{CrossAttn}(\mathbf{Q}_{net}, \mathbf{K}_{phr}, \mathbf{V}_{phr})\,
	\end{aligned}
	\label{eq:phrase Synergistic Utilization Mechanism}
\end{equation}
where, $\mathbf{K}_{phr}$ and $\mathbf{V}_{phr}$ are derived from $\mathbf{T}_{phr}$, $\mathbf{V}_{phr}^{'}\in\mathbb{R}^{L_{net}\times d_{net}}$. Eqs.~\ref{eq:global Synergistic Utilization Mechanism} and \ref{eq:phrase Synergistic Utilization Mechanism} share the $\mathbf{Q}_{net}$. The final output visual features of the target U-Net block are defined as:
\begin{equation}
	\begin{aligned}
		&\mathbf{\hat{V}}_{rsa} = \mathbf{V}_{glob}^{'} + \lambda\cdot \mathbf{V}_{phr}^{'}\ \in\mathbb{R}^{L_{net}\times d_{net}},\\
		&\lambda =
		\begin{cases}
			1, & \text{if in layout block}, \\
			0, & \text{otherwise.}
		\end{cases}
	\end{aligned}
	\label{eq:out Synergistic Utilization Mechanism}
\end{equation}
This design, as detailed in Section~\ref{sec:ablation_study}, achieves precise control over object quantities while minimizing compromises to the original text understanding and image generalization capabilities. The resulting framework not only enhances visual fidelity but also establishes a robust foundation for further fine-grained control extensions.
\subsection{Adaptive Multi-modal Guidance}
\label{sec:Adaptive Multi-modal Guidance}
As shown in Fig.~\ref{fig:framework}, the structure reference ($\mathbf{S}$), object reference ($\mathbf{O}$), and bounding box ($\mathbf{B}$) collectively form a multi-source control signal set $\mathbf{X}=\{\mathbf{S},\mathbf{O},\mathbf{B}\}$, where the activated control signals $\mathbf{C}$ serve as module input, with $\mathbf{C}\subseteq\mathbf{X}$. Since $\mathbf{S}$ and $\mathbf{B}$ serve as layout signals, they are not necessarily used simultaneously. Specifically, the AMG module comprises three components. Signal Encoder projects activated heterogeneous signals $\mathbf{C}$ into a unified feature representation $\mathbf{C}_{unif}\in\mathbb{R}^{L_{unif}\times d}$. Based on this representation, Adaptive Controller mines intrinsic correlations to formulate corresponding structured intent $\mathbf{C}_{str}\in\mathbb{R}^{L_{str}\times d}$. Finally, Intent-Guided Interaction Mechanism leverages $\mathbf{C}_{str}$ to dynamically modulate the spatial scope and attribute emphasis of constraints during generation, thereby achieving corresponding fine-grained control over scene layouts and object attributes.\\

\noindent\textbf{Signal Encoder.} 
To address the multi-source heterogeneity in input signals, we design the Signal Encoder for unified feature modeling. Taking the activated control signals $\mathbf{C}$ as input, the module performs deep extraction of multi-dimensional semantics and employs a self-attention mechanism to map heterogeneous features into a unified feature representation $\mathbf{C}_{unif}$. By establishing a semantically consistent feature space, the resulting representation eliminates modal bias arising from feature magnitude discrepancies. This unified formulation enables the subsequent Adaptive Controller to unbiasedly integrate all inputs, thereby accurately conveying structured intent $\mathbf{C}_{str}$.

Specifically, image-type signals are processed through a Dinov2 \cite{oquab2023dinov2} Image Encoder $\mathbf{E}_{{img}}$, while bounding boxes are modeled by a trainable Box Encoder $\mathbf{E}_{{box}}$. The resulting heterogeneous control features are then concatenated and aligned via a self-attention layer to form a unified feature representation $\mathbf{C}_{unif}$. This process can be formalized as: 
\begin{equation}
	\begin{aligned}
		\mathbf{C}_{unif} = \texttt{SelfAttn}([\mathbf{F}_s;\mathbf{F}_o;\mathbf{F}_b]),
	\end{aligned}
\end{equation}
where $[\ \cdot\ ;\ \cdot\ ]$ denotes concatenation. The $\mathbf{F}_i$ are defined as: 
\begin{equation}
    \mathbf{F}_i =
    \begin{cases}
    \mathcal{E}_i(\mathbf{X}_i), & \text{if } \mathbf{X}_i \in \mathbf{C}, \\
    \varnothing, & \text{otherwise},
    \end{cases}
    \quad i \in \{s, o, b\},
\end{equation}
where the inputs correspond to $\{\mathbf{X}_s, \mathbf{X}_o, \mathbf{X}_b\} = \{\mathbf{S}, \mathbf{O}, \mathbf{B}\}$. Here, $\mathcal{E}_i$ denotes the signal-specific encoder: specifically, $\mathcal{E}_s$ and $\mathcal{E}_o$ share $\mathbf{E}_{img}$, while $\mathcal{E}_b$ utilizes $\mathbf{E}_{box}$. Regarding the dimensions, we have $\mathbf{F}_s \in \mathbb{R}^{L_{s}\times d}$, $\mathbf{F}_o \in \mathbb{R}^{L_{o}\times d}$, and $\mathbf{F}_b \in \mathbb{R}^{L_{b}\times d}$. When $\mathbf{F}_i = \varnothing$, we set the dimension to $\mathbb{R}^{0 \times d}$. The total sequence length is given by $L_{unif} = L_{s} + L_{o} + L_{b}$.\\

% where the inputs correspond to $\{\mathbf{X}_s, \mathbf{X}_o, \mathbf{X}_b\}=\{\mathbf{S},\mathbf{O}, \mathbf{B}\}$, $\mathcal{E}_i$ denotes the signal-specific encoder, $\mathcal{E}_s$ and $\mathcal{E}_o$ share the image encoder $\mathbf{E}_{img}$, $\mathcal{E}_b$ utilizes the box encoder $\mathbf{E}_{box}$,
% $\mathbf{F}_s\in\mathbb{R}^{L_{s}\times d}$, $\mathbf{F}_o\in\mathbb{R}^{L_{o}\times d}$, $\mathbf{F}_b\in\mathbb{R}^{L_{b}\times d}$. If $\mathbf{F}_i=\varnothing$, we set $\mathbf{F}_i\in\mathbb{R}^{0\times d}$. The total sequence length $L_{unif}$ of $\mathbf{C}_{unif}$ is defined as $L_{unif}=L_{s}+L_{o}+L_{b}$.

\noindent\textbf{Adaptive Controller.} 
This component is designed to transform the unified feature representation $\mathbf{C}_{unif}$ into the structured intent $\mathbf{C}_{str}$ via a dual-stage attention mechanism. Diverging from conventional static condition aggregation, we employ a dynamic interaction paradigm to accommodate the diverse signal combinations within $\mathbf{C}_{unif}$. Specifically, learnable query $\mathbf{Q}_{str}\in\mathbb{R}^{L_{str}\times d}$ first utilize cross-attention to adaptively aggregate contained signal semantics from $\mathbf{C}_{unif}$, and are then processed by a self-attention mechanism to model global dependencies among the integrated semantics. This design resolves potential ambiguities across multimodal signals and precisely binds disjoint constraints, such as spatial bounding boxes and specific attribute descriptions, thereby yielding structured intent $\mathbf{C}_{str}$ that defines spatial layout and visual attribute specifications of constraints. Consequently, $\mathbf{C}_{str}$ provides the generative backbone with semantically comprehensive and explicit control instructions. This process can be formalized as:
\begin{equation}
	\begin{aligned}
        &\mathbf{C}_{unif}^{'} = \texttt{SelfAttn}(\texttt{CrossAttn}(\mathbf{Q}_{str}, \mathbf{K}_{unif}, \mathbf{V}_{unif})),\\
		&\mathbf{C}_{str} = \texttt{FFN}(\texttt{LN}(\mathbf{C}_{unif}^{'})),\\
	\end{aligned}
\end{equation}
where, $\mathbf{C}_{unif}^{'}\in\mathbb{R}^{L_{str}\times d}$ denotes the intermediate variable, $\mathbf{K}_{unif}$ and $\mathbf{V}_{unif}$ are derived from $\mathbf{C}_{unif}$.\\
\begin{figure}[!t]
	\centering
	\begin{overpic}[width=0.5\textwidth, trim=18 0 0 18, clip]{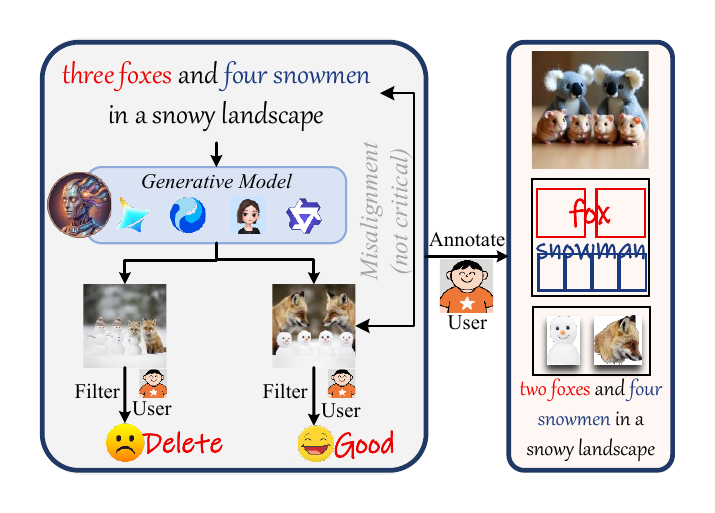}
		%\put(23,13){\parbox{4cm}{$t_g$}}
		%\put(60,13.5){\parbox{4cm}{$t_p$ ($\varnothing $)}}
	\end{overpic}
	\vspace{-25pt}
	\caption{\textbf{The construction pipeline of the MoCA benchmark.} First, diverse predefined text inputs are used with generative models to produce candidate images. Next, manual screening retains high-quality images with clear object attributes, without requiring quantity consistency between the generated results and the input text. Finally, selected images are manually annotated to establish a multi-dimensional benchmark comprising text descriptions, structure references, object references, and bounding boxes.}
	\label{fig:dataset}
	\vspace{0pt}
\end{figure}
\begin{table*}[!t]
	\centering
	\caption{\textbf{Quantitative comparisons under different generation configurations}. The \textcolor{red}{best} and \textcolor{blue}{second best} have been marked by red and blue, respectively. “N/A” represents not applicable.}
	\label{tab:comparisons}
	\vspace{-5pt}
	\renewcommand{\arraystretch}{1.1}
	\resizebox{\textwidth}{!}{
		\begin{threeparttable}
			\begin{tabular}{ccccccccccc}
				\hline
				\multirow{2}{*}{\textbf{Method}} &   & \multirow{2}{*}{\textbf{NIQE ↓}} & \multirow{2}{*}{\textbf{CLIP Score ↑}} & \multirow{2}{*}{\textbf{DPG Score ↑}} & \multicolumn{2}{c}{\textbf{Q-Align}} & \multirow{2}{*}{\textbf{IP-Sim↑}} & \multirow{2}{*}{\textbf{Spatial-Sim ↑}} & \multirow{2}{*}{\textbf{Numerical ↑}} & \multirow{2}{*}{\textbf{MOS ↑}} \\ \cline{6-7}
				& & & & & \textbf{Quality ↑} & \textbf{Aesthetic ↑} & & & & \\ 
				\hline

				\multicolumn{11}{c}{\lga\textit{\textbf{Text-to-image generation (T → Image)}}} \\
				\hline
				% -------- Block 1: T ↓ Image --------
				\textbf{SDXL} &  \cite{podell2023sdxl} &  2.817 &  0.292 &  84.26 &  4.715 &  4.147 &  N/A &  N/A &  8.72 &  3.96 \\
				\textbf{Emu2} & \cite{wang2024genartist} & 2.991 & 0.297 & 83.77 & 4.491 & 4.089 & N/A & N/A & 6.72 & 2.31 \\
				\textbf{Omnigen2} &  \cite{wu2025omnigen2} &  2.687 &  0.301 &  \textcolor{blue}{{89.91}} &  4.745 &  4.250 &  N/A &  N/A &  12.72 &  18.96 \\
				\textbf{FLUX} & \cite{lan2025flux} & \textcolor{blue}{{2.675}} & \textcolor{blue}{{0.337}} & 88.05 & \textcolor{red}{{4.901}} & \textcolor{blue}{4.311} & N/A & N/A & \textcolor{blue}{{15.73}} & \textcolor{blue}{{27.21}} \\
				\lb \textbf{MoGen (Ours)} & \lb  & \lb \textcolor{red}{{2.607}} & \lb \textcolor{red}{{0.342}} & \lb \textcolor{red}{{92.47}} & \lb \textcolor{blue}{{4.857}} & \lb \textcolor{red}{{4.492}} & \lb N/A & \lb N/A & \lb \textcolor{red}{{65.28}} & \lb \textcolor{red}{{67.75}} \\

				\hline
				\multicolumn{11}{c}{\lga\textit{\textbf{Text+object to image generation (T+O → Image)}}} \\
				\hline
				% -------- Block 2: T+O ↓ Image --------
				\textbf{MS-diffusion} & \cite{wangms} & 2.672 & 0.306 & 85.38 & 4.582 & 4.129 & 0.654 & N/A & 11.72 & 1.36 \\
				\textbf{Omnigen2} &  \cite{wu2025omnigen2} &  \textcolor{blue}{{2.610}} &  \textcolor{blue}{{0.332}} &  90.82 &  \textcolor{red}{{4.891}} &  \textcolor{red}{{4.517}} &  0.697 &  N/A &  \textcolor{blue}{{33.81}} &  \textcolor{blue}{{8.94}} \\
				\textbf{Xverse} & \cite{chen2025xverse} & 2.624 & 0.328 & \textcolor{blue}{{91.35}} & 4.702 & 4.376 & \textcolor{blue}{{0.705}} & N/A & 21.64 & 5.72 \\
				\lb \textbf{MoGen (Ours)} & \lb  & \lb \textcolor{red}{{2.575}} & \lb \textcolor{red}{{0.349}} & \lb \textcolor{red}{{93.01}} & \lb \textcolor{blue}{{4.801}} & \lb \textcolor{blue}{{4.448}} & \lb \textcolor{red}{{0.762}} & \lb N/A & \lb \textcolor{red}{{68.91}} & \lb \textcolor{red}{{83.98}} \\

				\hline
				\multicolumn{11}{c}{\lga\textit{\textbf{Text+box to image generation (T+B → Image)}}} \\
				\hline
				% -------- Block 3: T+B ↓ Image --------
				\textbf{Bounded-attention} & \cite{dahary2024yourself} & 3.172 & \textcolor{blue}{{0.316}} & \textcolor{blue}{{86.42}} & 4.471 & 3.989 & N/A & \textcolor{blue}{{0.557}} & \textcolor{blue}{{31.14}} & \textcolor{blue}{{3.81}} \\
				\textbf{Emu2} &  \cite{wang2024genartist} &  \textcolor{blue}{{2.921}} &  0.307 &  85.29 &  \textcolor{blue}{{4.515}} &  \textcolor{blue}{{4.106}} &  N/A &  0.472 &  25.49 &  0.57 \\
				\lb \textbf{MoGen (Ours)} & \lb & \lb\textcolor{red}{{2.635}} & \lb\textcolor{red}{{0.334}} & \lb\textcolor{red}{{92.73}} & \lb\textcolor{red}{{4.725}} & \lb\textcolor{red}{{4.429}} & \lb N/A & \lb\textcolor{red}{{0.701}} & \lb\textcolor{red}{{74.91}} & \lb\textcolor{red}{{95.62}} \\

				\hline
				\multicolumn{11}{c}{\lga\textit{\textbf{Text+object+box to image generation (T+O+B → Image)}}} \\
				\hline
				% -------- Block 4: T+O+B ↓ Image --------
				\textbf{Emu2} &  \cite{wang2024genartist} &  2.916 &  0.327 &  85.73 &  4.479 &  4.192 &  0.662 &  0.458 &  23.81 &  1.32 \\
				\textbf{MS-diffusion} & \cite{wangms} & \textcolor{blue}{{2.797}} & \textcolor{blue}{{0.336}} & \textcolor{blue}{{87.05}} & \textcolor{blue}{{4.531}} & \textcolor{blue}{{4.214}} & \textcolor{blue}{{0.671}} & \textcolor{blue}{{0.503}} & \textcolor{blue}{{26.29}} & \textcolor{blue}{{7.16}} \\
				\lb \textbf{MoGen (Ours)} & \lb  & \lb \textcolor{red}{{2.593}} & \lb \textcolor{red}{{0.351}} & \lb \textcolor{red}{{92.95}} & \lb \textcolor{red}{{4.785}} & \lb \textcolor{red}{{4.514}} & \lb \textcolor{red}{{0.723}} & \lb \textcolor{red}{{0.675}} & \lb \textcolor{red}{{76.26}} & \lb \textcolor{red}{{91.52}} \\

				\hline
			\end{tabular}
		\end{threeparttable}
	}
\end{table*}

\begin{table*}[!t]
	\centering
	\caption{\textbf{Quantitative comparisons under text+structure (\textbf{\textit{T+S~→~Image}}) and text+structure+object (\textbf{\textit{T+S+O~→~Image}}) configurations}. The \textcolor{red}{best} and \textcolor{blue}{second best} have been marked by red and blue, respectively. “N/A” represents not applicable.}
	\label{tab:comparison-T+S+O}
	\vspace{-5pt}
	\renewcommand{\arraystretch}{1.1}
	\resizebox{\textwidth}{!}{
		\begin{threeparttable}
			\begin{tabular}{cccccccccccc}
				\hline
				\multirow{2}{*}{\textbf{Method}} &   & \multirow{2}{*}{\textbf{NIQE$\downarrow$}} & \multirow{2}{*}{\textbf{CLIP Score$\uparrow$}} & \multirow{2}{*}{\textbf{DPG Score$\uparrow$}} & \multicolumn{2}{c}{\textbf{Q-Align}} & \multicolumn{2}{c}{\textbf{Similarity}} & \multirow{2}{*}{\textbf{Spatial-Sim$\uparrow$}} & \multirow{2}{*}{\textbf{Numerical$\uparrow$}} & \multirow{2}{*}{\textbf{MOS$\uparrow$}} \\ \cline{6-9}
				&   &                       &                             &                            & \textbf{Quality$\uparrow$}      & \textbf{Aesthetic$\uparrow$}    & \textbf{IP-Sim$\uparrow$}         & \textbf{Img-Sim$\downarrow$}       &                              &                            &                      \\ \hline
				\textbf{StableFlow}              & \cite{avrahami2025stable} &  2.707                 & .297                       & 87.72                      & 4.658        & 4.372        &  N/A              & 0.796         & 0.681                        & 43.01                      & 3.78                 \\
				\textbf{FlowEdit}                & \cite{kulikov2024flowedit} & 2.618                 & 0.314                       & 89.57                      & 4.683        & 4.471        & N/A              & 0.757         & \textcolor{red}{{0.742}}      & 61.72                      & 15.77                \\
				\textbf{MoEdit}                  & \cite{li2025moedit} & 2.748                 & 0.323                       & 90.37                      & 4.702        & 4.426        &  N/A              & 0.764         & 0.672                        & 54.92                      & 21.76                \\
				\lgr\textbf{MoGen (T+S)}             & \lgr  & \lgr\textcolor{blue}{{2.573}}        & \lgr\textcolor{blue}{{0.337}}              & \lgr\textcolor{blue}{{93.07}}             & \lgr\textcolor{red}{{4.795}} & \lgr\textcolor{blue}{{4.513}} & \lgr N/A              & \lgr\textcolor{blue}{{0.632}} & \lgr0.685                        & \lgr\textcolor{blue}{{78.72}}             & \lgr\textcolor{red}{{31.20}}       \\
				\lb\textbf{MoGen (T+S+O)}           & \lb  & \lb\textcolor{red}{{2.553}}        & \lb\textcolor{red}{{0.342}}              & \lb\textcolor{red}{{93.42}}             & \lb\textcolor{blue}{{4.737}} & \lb\textcolor{red}{{4.526}} & \lb\textcolor{red}{{0.731}} & \lb\textcolor{red}{{0.604}} & \lb\textcolor{blue}{{0.691}}              & \lb\textcolor{red}{{79.93}}             & \lb\textcolor{blue}{{27.49}}       \\ \hline
			\end{tabular}
		\end{threeparttable}
	}
\end{table*}

\noindent\textbf{Intent-Guided Interaction Mechanism.}
This mechanism is introduced to bridge the semantic gap between high-level intents and low-level generative features. It comprises learnable interaction layers that act as semantic adapters and are designed to precisely map the structured intent $\mathbf{C}_{str}$ from the Adaptive Controller into the feature space of the U-Net. Through this mapping, the $\mathbf{C}_{str}$ is transformed into control instructions that are interpretable and executable by the U-Net, allowing for the modulation of spatial constraints and attribute emphasis during generation. Specifically, as illustrated in Fig.~\ref{fig:framework}, each interaction layer is instantiated as a cross-attention module. It utilizes pre-trained U-Net query to retrieve control semantic cues from $\mathbf{C}_{str}$ for feature space alignment. The computation is formalized as follows:
\begin{equation}
	\begin{aligned}
		\mathbf{\hat{V}}_{amg} = \texttt{CrossAttn}(\mathbf{Q}_{net}, \mathbf{K}_{str}, \mathbf{V}_{str}),
	\end{aligned}
	\label{eq:Dynamic Scheduling AMG}
\end{equation}
where $\mathbf{K}_{str}$ and $\mathbf{V}_{str}$ are derived from $\mathbf{C}_{str}$, $\mathbf{Q}_{net}$ denotes the query from Eqs.~\ref{eq:global Synergistic Utilization Mechanism} and \ref{eq:phrase Synergistic Utilization Mechanism}, and $\mathbf{\hat{V}}_{amg}\in\mathbb{R}^{L_{net}\times d_{net}}$ represents the aligned control features. To impose fine-grained constraints, we inject these instructions into the generation stream via a residual connection: $\mathbf{\hat{V}}_{mogen} = \mathbf{\hat{V}}_{amg} + \mathbf{\hat{V}}_{rsa}$, where $\mathbf{\hat{V}}_{rsa}$ corresponds to the output from Eq.~\ref{eq:out Synergistic Utilization Mechanism}, and $\mathbf{\hat{V}}_{mogen}\in\mathbb{R}^{L_{net}\times d_{net}}$ signifies the final feature representation integrated with control signals.
\section{Multi-Object Control-aware Annotation}
\label{sec:MoCA}
Multi-dimensional annotations play a crucial role for multi-object image generation. However, the field currently lacks publicly available benchmarks with sufficient data and multi-dimensional annotations, as the annotation complexity scales significantly with the number of objects per image. This gap has become a critical bottleneck hindering further research. To address this challenge, we construct a multi-object image benchmark with multi-dimensional annotations, providing essential data resources to advance research in this field and establishing a solid experimental foundation for our proposed method.

MoCA contains over $10K$ multi-object images synthesized by existing generative models \cite{lan2025flux, li2024hunyuan}. As shown in Fig.~\ref{fig:dataset}, we first generate candidate samples at scale based on predefined diverse text inputs, then manually filter to retain high-quality images with clear object attributes, without requiring strict correspondence between object quantity and input text. We further construct a multi-dimensional annotation system for the filtered images to ensure data diversity and representativeness. The system encompasses four core annotation types: text descriptions provide semantic-level natural language expressions, structure references depict overall scene layouts, object references capture fine-grained visual attributes of instances, and bounding boxes enable precise target localization. For annotation specifications, each image contains $1$-–$15$ objects and is equipped with one text description and one structure reference. Each object includes a bounding box and an object reference, forming a multi-level annotation system spanning from global semantics to local appearance.
\begin{figure*}[!t]
	\centering
	\begin{overpic}[width=1.0\textwidth, trim=18 0 20 15, clip]{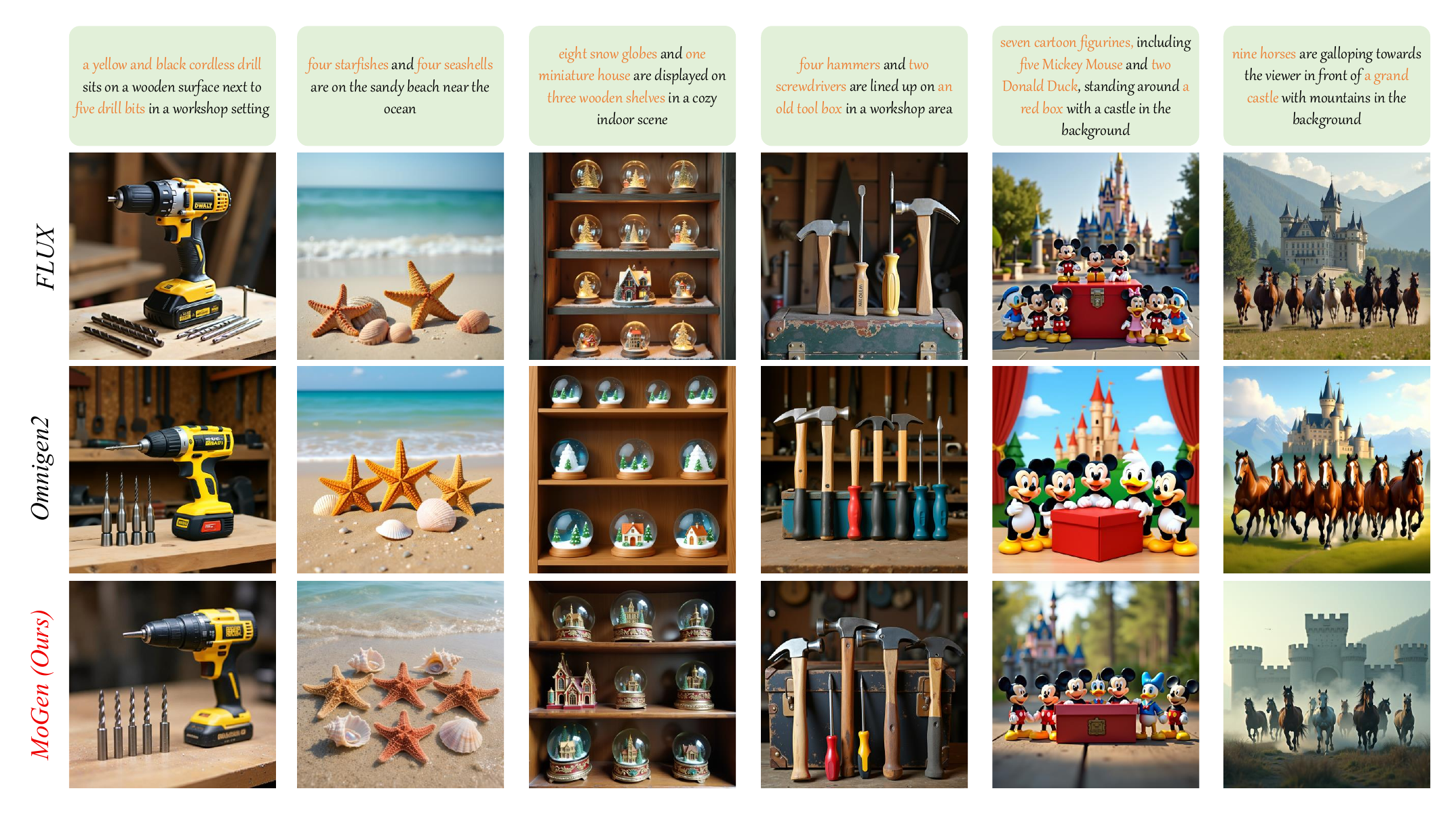}
		
	\end{overpic}
	\vspace{-25pt}
	\caption{\textbf{Qualitative comparisons on \textit{\textbf{T~→~Image}}}. MoGen achieves superior performance in text-to-multi-object image generation, excelling in quantity consistency, image quality, and semantic alignment. FLUX \cite{flux2024} ranks second with partial limitations, Omnigen2 \cite{wu2025omnigen2} shows moderate accuracy.}
	\label{fig:comparison-text}
	\vspace{-5pt}
\end{figure*}
\section{Experiment}
\subsection{Implementation}
\noindent\textbf{Training Details.} The experiments are conducted using the PyTorch framework on $8$ NVIDIA H20 GPUs, with SDXL~\cite{podell2023sdxl} serving as the baseline model. The training process comprises two stages. In the first stage, only the RSA module and its corresponding U-Net interaction layer parameters are updated. In the second stage, the weights obtained from the first stage are frozen, and only the AMG module and its corresponding U-Net interaction layer parameters are trained. The training dataset is partitioned into two subsets corresponding to these two stages. We employ the Adam optimizer for over $45K$ training steps, with the learning rate gradually decaying from $5\times10^{-5}$ to $5\times10^{-6}$. All inference is performed on a single H20 GPU.
\begin{figure*}[t]
	\centering
	\begin{overpic}[width=1.0\textwidth, trim=18 0 20 15, clip]{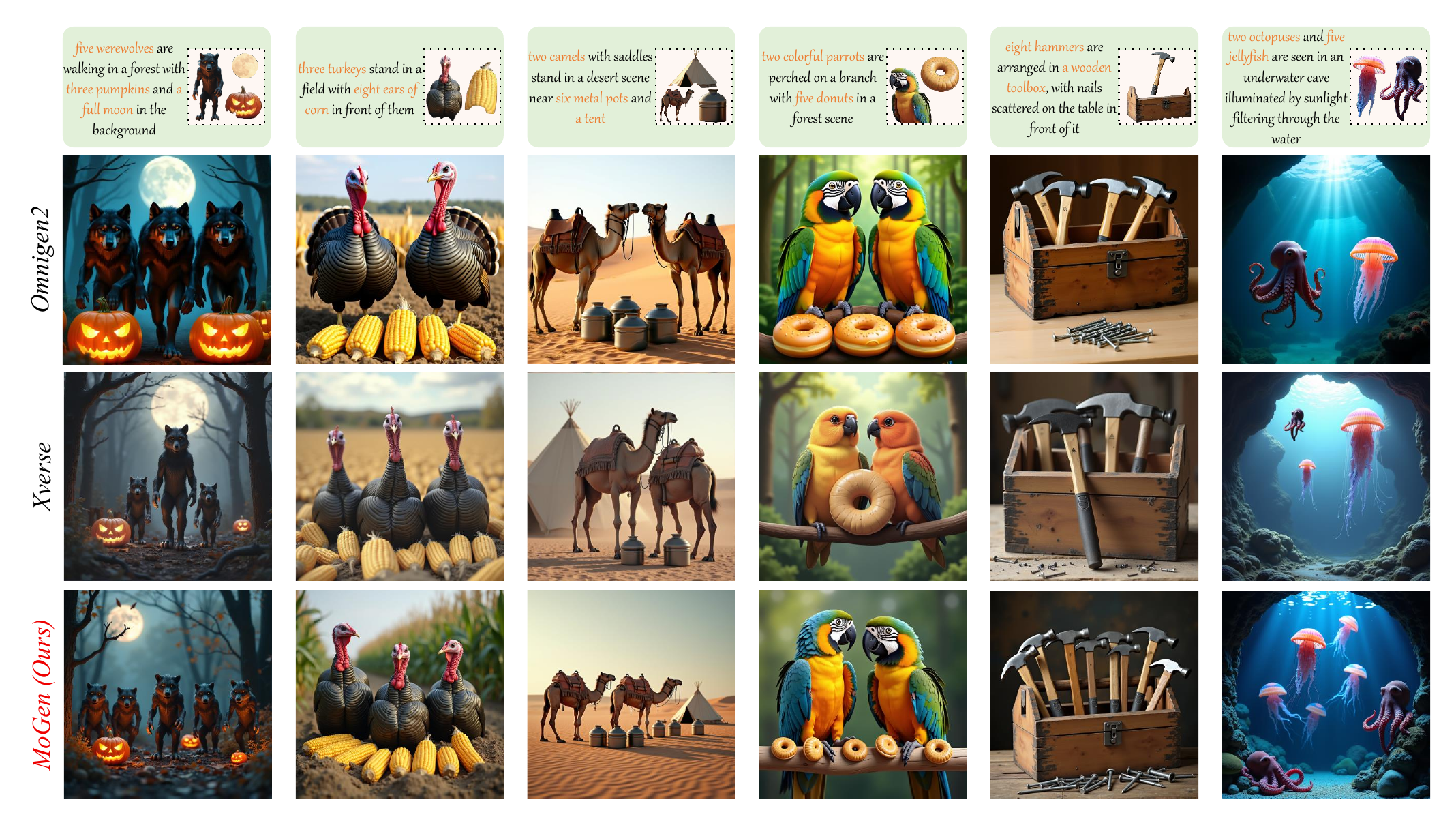}
		
	\end{overpic}
	\vspace{-25pt}
	\caption{\textbf{Qualitative comparisons on \textit{\textbf{T+O~→~Image}}}. MoGen ensures quantity consistency, faithful appearance preservation, and coherent scene layouts, leading to superior realism. In comparison, Omnigen2 \cite{wu2025omnigen2} achieves good aesthetics but lacks quantity consistency, Xverse \cite{chen2025xverse} restores textures well but often omits or duplicates objects.}
	\label{fig:comparison-text+obj}
	\vspace{0pt}
\end{figure*}
\begin{figure*}[!t]
	\centering
	\begin{overpic}[width=1.0\textwidth, trim=18 0 20 0, clip]{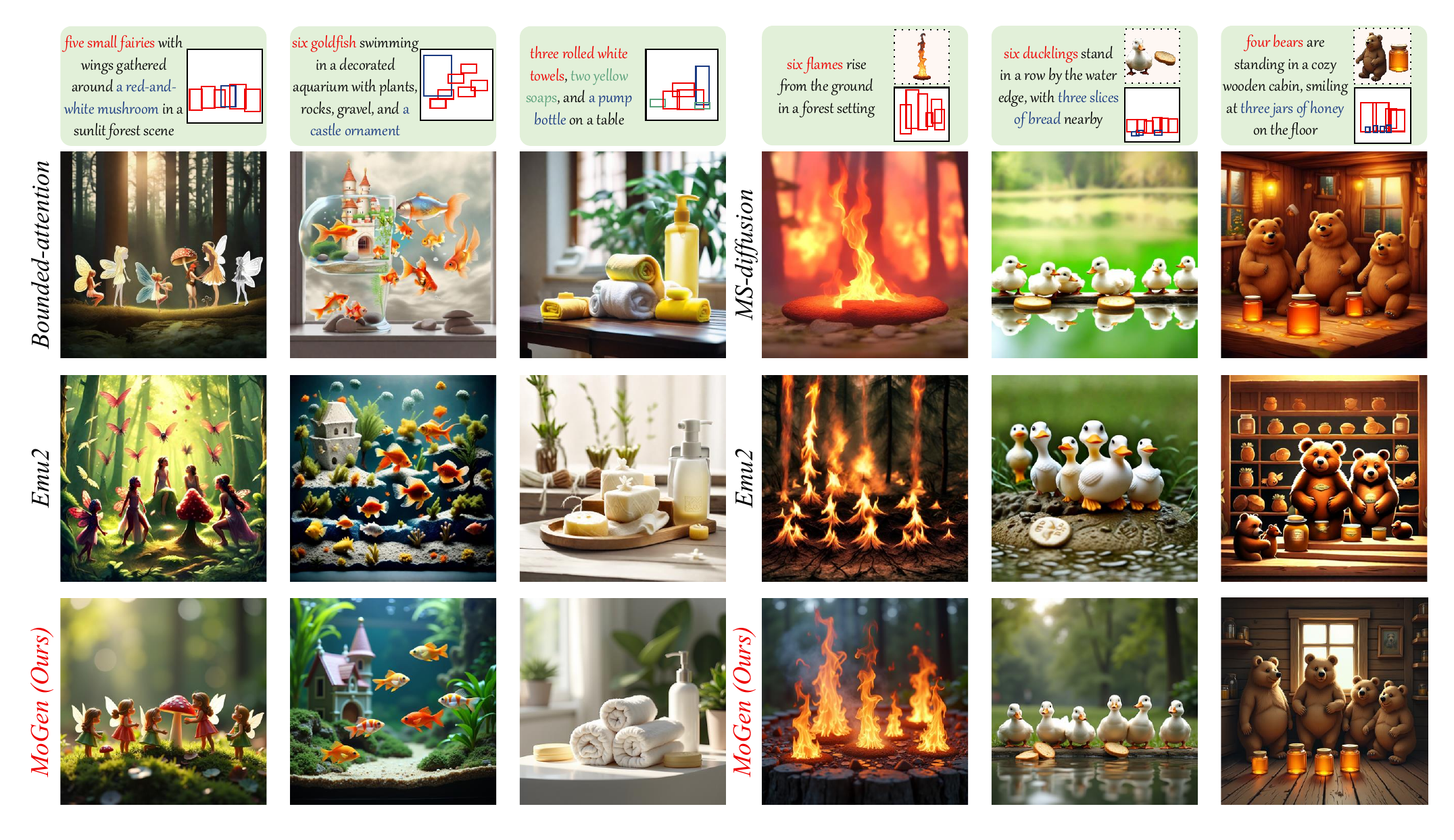}
		\put(0,0){\parbox{4cm}{(a) \textit{\textbf{T+B~→~Image}}}}
        \put(50.5,0){\parbox{4cm}{(b) \textit{\textbf{T+B+O~→~Image}}}}
	\end{overpic}
	\vspace{-15pt}
	\caption{\textbf{Qualitative comparisons on \textit{\textbf{T+B~→~Image}} and \textit{\textbf{T+B+O~→~Image}}}. MoGen achieves superior performance in quantity consistency, spatial alignment, and visual fidelity. It strictly adheres to textual and bounding box constraints while preserving the morphological and appearance details of object references. In contrast, Bounded-attention \cite{dahary2024yourself} and MS-diffusion \cite{wangms} suffer from spatial and quantity deviations, whereas Emu2 \cite{sun2024generative} often fails to control object quantity and distribution despite producing natural visual results.}
	\label{fig:comparison-text+box}
	\vspace{0pt}
\end{figure*}
% \begin{figure*}[!t]
% 	\centering
% 	\begin{overpic}[width=1.0\textwidth, trim=18 0 20 18, clip]{fig/comparison-box+obj.pdf}
		
% 	\end{overpic}
% 	\vspace{-25pt}
% 	\caption{\textbf{Qualitative comparisons on T+O+B~$\rightarrow$~Image}. MoGen achieves superior performance in quantity consistency by precisely following textual specifications, spatial alignment through accurate object-to-bounding-box correspondences, and reference image fidelity by maintaining morphological and textural characteristics compared to MS-diffusion \cite{wangms} and Emu2 \cite{sun2024generative}, demonstrating enhanced controllability and stability.}
% 	\label{fig:comparison-text+box+obj}
% 	\vspace{0pt}
% \end{figure*}
\begin{figure*}[!t]
	\centering
	\begin{overpic}[width=1.0\textwidth, trim=18 0 20 18, clip]{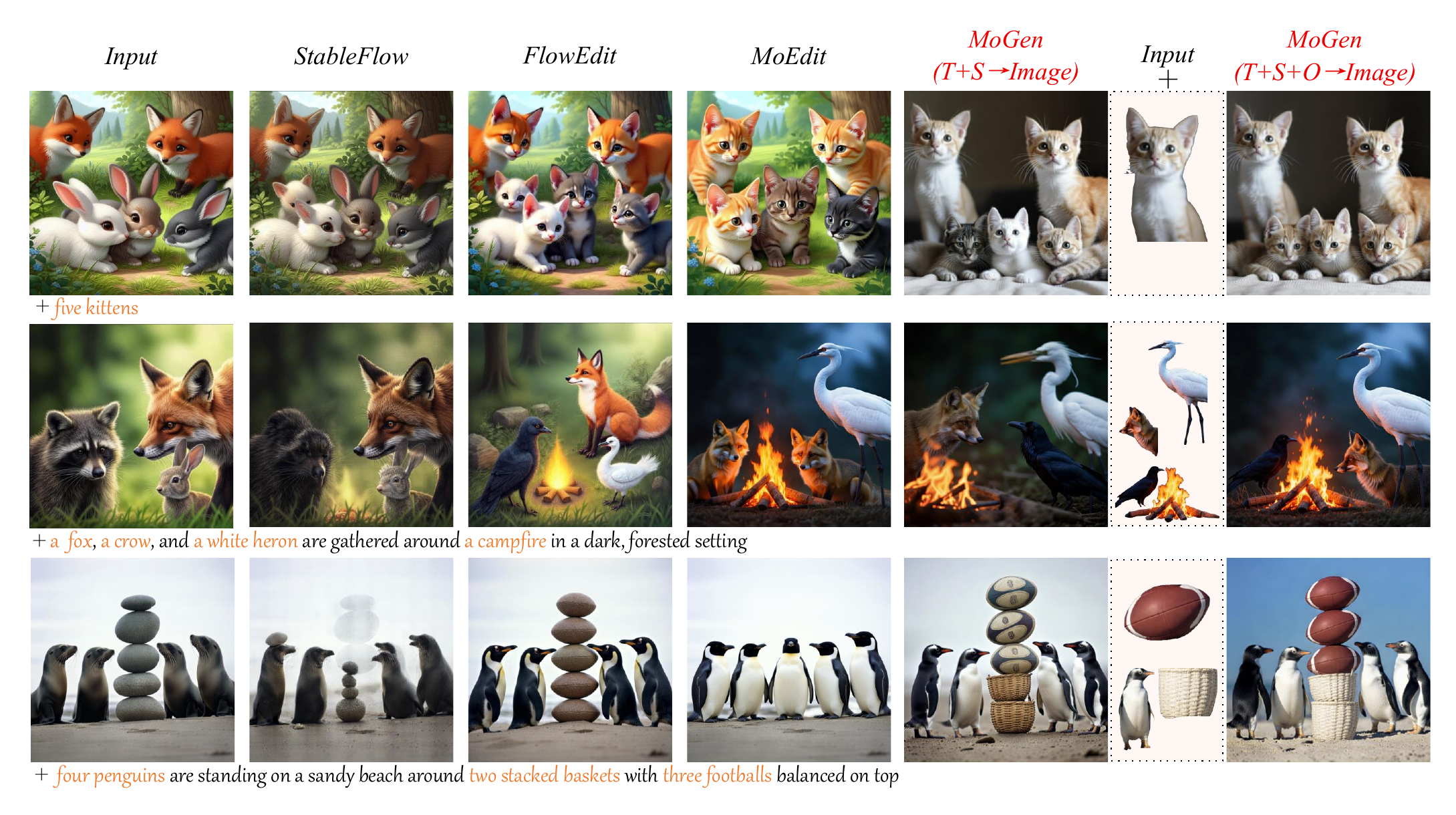}
		
	\end{overpic}
	\vspace{-25pt}
	\caption{\textbf{Qualitative comparisons on \textit{\textbf{T+S~→~Image}} and \textit{\textbf{T+S+O~→~Image}}}. MoGen achieves quantity and spatial consistency. It demonstrates the highest diversity in appearance and creativity, with MoGen (T+S+O) showing greater fidelity to object appearances. In contrast, MoEdit \cite{li2025moedit} exhibits limitations in multi-category tasks, StableFlow \cite{avrahami2025stable} suffers from quantity deviations, and FlowEdit \cite{kulikov2024flowedit} shows unstable structure preservation.}
	\label{fig:comparison-text+str+obj}
	\vspace{0pt}
\end{figure*}

\noindent\textbf{Dataset.} Model training employs the MoCA dataset constructed in this work. To enhance model robustness, we apply four data augmentation techniques: (1) Bounding Box Jittering: We apply slight random perturbations to target bounding boxes to improve tolerance to localization errors. (2) Local Quality Distortion: Local regions in input images are randomly selected and subjected to geometric distortion or quality degradation operations. (3) Random Cropping for Reference Images: Object references undergo random cropping with cropping ratios not exceeding $15\%$ of the original image area, ensuring sufficient discriminative information is preserved while introducing appearance variations.
%(4) Background Handling: For background information in appearance references, we employ two processing strategies: replacing background regions with pure white to eliminate interference, and preserving partial background regions inevitably captured during bounding box extraction to improve adaptability in real applications.
\subsection{Quantitative Comparisons}
In this study, we select $9$ models as baseline methods, including SDXL \cite{podell2023sdxl}, FLUX \cite{flux2024}, Emu2 \cite{sun2024generative}, Omigen2 \cite{wu2025omnigen2}, MS-diffusion \cite{wangms}, Xverse \cite{chen2025xverse}, Bounded-attention \cite{dahary2024yourself}, StableFlow \cite{avrahami2025stable}, and FlowEdit \cite{kulikov2024flowedit} (see Tab.~\ref{tab:previous_method}).
To comprehensively evaluate model performance, we employ $9$ metrics: NIQE \cite{mittal2012making}, CLIP Score \cite{radford2021learning}, DPG Score \cite{hu2024ella}, Q-Align \cite{wu2023q}, IP-Sim \cite{radford2021learning}, Spatial-Sim, Img-Sim \cite{radford2021learning}, Numerical Accuracy, and MOS. NIQE measures the natural perceptual quality of images. CLIP Score quantifies semantic alignment by computing CLIP similarity between generated images and text prompts. DPG Score evaluates proposition-level constraint satisfaction in generated images, capturing fine-grained consistency of object attributes and relationships. Q-Align assesses compositional and aesthetic quality. IP-Sim measures appearance similarity between generated targets and input reference images. Spatial-Sim evaluates consistency between generated instance layouts and structure references or bounding boxes. Img-Sim quantifies perceptual similarity between image pairs. Numerical Accuracy examines adherence to quantity constraints in generation results. Finally, MOS reflects subjective perceptual quality as rated by users. To ensure objective and reliable evaluation, we recruit $12$ participants for subjective assessment.

As shown in Table~\ref{tab:comparisons}, in text-to-image (\textit{\textbf{T~→~Image}}) scenario, MoGen ensures robust image quality while marking a significant advancement in semantic and quantity consistency. Beyond achieving relative gains in NIQE ($2.5\%$), CLIP Score ($1.5\%$), and DPG Score ($2.8\%$), MoGen exceeds the leading baseline by $315\%$ in quantity consistency (Numerical), a breakthrough that propels the MOS to $67.75$. While backbone disparities between SDXL and FLUX cause a slight $0.9\%$ regression in Q-Align, MoGen effectively mitigates these limitations, delivering markedly superior aesthetic quality and user satisfaction.

In text+object to image (\textit{\textbf{T+O~→~Image}}), MoGen continuously demonstrates superiority in object consistency and conditional adherence. It achieves improvements of $8.1\%$, $5.1\%$, and $1.8\%$ in IP-Sim, CLIP Score, and DPG Score, respectively, while outperforming baselines in quantity consistency (Numerical) by approximately $104\%$. Despite slight $1–2\%$ of Omnigen2 lead in Q-Align, achieved by compromising object fidelity (IP-Sim), MoGen ranks first in human evaluation with a MOS of $83.98$, validating its comprehensive advantage in multi-object generation scenarios.

In text+box to image (\textit{\textbf{T+B~→~Image}}), MoGen effectively balances visual quality with layout constraints. Specifically, improvements of $9.8\%$, $25.9\%$, and $141\%$ in NIQE, Spatial-Sim, and Numerical, respectively, drive MOS to a notable $95.62$. These results demonstrate robustness in layout-content coupling; moreover, concurrent gains in CLIP and DPG scores confirm that the method enhances controllability without compromising textual semantic consistency.

In text+object+box to image (\textit{\textbf{T+O+B~→~Image}}), MoGen establishes a consistent lead across all metrics. It improves Spatial-Sim, Numerical, and IP-Sim by $34.2\%$, $190\%$, and $7.7\%$, while leading in generation quality and semantic alignment metrics (NIQE, CLIP, DPG, and Q-Align). A MOS of $91.52$ further underscores its competitive edge in human alignment. Overall, these results strongly attest to the comprehensive, end-to-end improvement in capabilities of MoGen for complex controllable generation.

Table~\ref{tab:comparison-T+S+O} compares the generative performance under text+structure to image (\textit{\textbf{T+S~→~Image}}). We introduce FlowEdit and StableFlow as baselines and additionally employ the Image-Sim metric to assess input-output image similarity, where lower values are preferred to overcome the strong anchoring effects of reference images—unlike editing tasks that prioritize similarity preservation.
We treat structure references as hierarchical constraints that provide richer layout information than detection boxes, enabling stable layout control particularly in multi-object crossing and occlusion scenarios while preserving generation flexibility. Since existing methods cannot simultaneously incorporate multi-object object references, we compare MoGen under the text+structure+object to image (\textit{\textbf{T+S+O~→~Image}}) against \textit{\textbf{T+S~→~Image}} baselines, which remain representative for structural preservation despite different conditions.

Under \textit{\textbf{T+S~→~Image}}, MoGen excels in semantic alignment, boosting CLIP and DPG scores by $4.3\%$ and $3.0\%$, respectively. It also ensures robust perceptual quality (NIQE $1.7\%$ reduction) and a substantial gain in quantity consistency ($27.5\%$ increase). Although Spatial-Sim trails FlowEdit by $7.7\%$, this reflects a strategic trade-off: unlike editing-based methods that sacrifice expressive freedom (Img-Sim) for spatial rigidity, MoGen prioritizes generative diversity. Additionally, Spatial-Sim is biased as it depends on successful object generation.
The \textit{\textbf{T+S+O~→~Image}} further enforces instruction following, lifting CLIP, DPG, and Numerical metrics by $1.5\%$, $0.4\%$, and $1.5\%$ over the T+S baseline. While Img-Sim rises by $4.4\%$ due to rigid attribute constraints, this prioritization of fidelity over aesthetics leads to an $11.9\%$ drop in MOS. Despite Spatial-Sim remaining $6.9\%$ below FlowEdit, MoGen demonstrates superior overall balance and robustness compared to state-of-the-art methods.

Overall, MoGen demonstrates consistent and substantial advantages across diverse generation protocols. This comprehensive enhancement across multiple evaluation dimensions validates the effectiveness and robustness of MoGen in multi-object image generation tasks. As an extension of the prior MoEdit framework, this work effectively addresses key limitations including insufficient multi-category editing capabilities, constrained expression freedom due to visual anchor dependencies, and restrictive single-configuration constraints, thereby establishing a more generalizable and efficient generation paradigm.
\subsection{ Qualitative comparisons.}
\noindent\textit{\textbf{T~→~Image.}} Fig.~\ref{fig:comparison-text} presents a visual quality comparison of different methods for text-to-multi-object image generation. %Overall, MoGen demonstrates superior performance across three key evaluation dimensions: quantity consistency, image quality, and semantic alignment. 
Specifically, MoGen consistently generates target instances that closely match input descriptions, achieving significantly better quantity consistency, image clarity, and detail preservation compared to competing approaches. The method also exhibits enhanced semantic coherence in terms of object categories, spatial layout, and background composition.
FLUX ranks second in overall performance, showing comparable but imperfect results in quantity control and semantic alignment. Omnigen2 demonstrates moderate performance with notable deficiencies in quantity accuracy and detail rendering. 
%SDXL and Emu2 exhibit relatively weaker performance. While SDXL maintains reasonable visual quality, it suffers from quantity deviations and semantic confusion. Emu2 performs poorly across all three dimensions, producing results with evident blur, missing objects, and semantic errors.
In summary, MoGen achieves optimal performance, substantially advancing the state-of-the-art in text-to-multi-object image generation.

\noindent\textit{\textbf{T+O~→~Image.}} As shown in Fig.~\ref{fig:comparison-text+obj}, MoGen demonstrates superior performance in both consistency and realism of generated results. The generated scenes effectively follow quantity requirements specified in input descriptions while maintaining faithful reproduction of reference appearances. Additionally, MoGen exhibits high coherence and naturalness in scene layout and detail rendering, achieving enhanced visual realism.
In contrast, Omnigen2 shows satisfactory appearance restoration and aesthetic quality in certain scenarios but lacks strict quantity control, resulting in deviations from target descriptions. Xverse exhibits advantages in object texture and detail restoration but suffers from insufficient quantity consistency, frequently producing omissions or duplications. 
%MS-diffusion tends toward stylized content generation and, while visually appealing in isolated cases, demonstrates limited object completeness and semantic consistency, making it inadequate for complex multi-object scenarios.
In summary, MoGen achieves an optimal balance among quantity consistency, appearance preservation, and overall aesthetic quality, significantly outperforming existing approaches in visual generation tasks.

\noindent\textit{\textbf{T+B~→~Image}} and \textit{\textbf{T+B+O~→~Image}.} As illustrated in Fig.~\ref{fig:comparison-text+box}, MoGen demonstrates superior performance across quantity consistency, spatial alignment, and visual fidelity, significantly outperforming Bounded-attention, MS-diffusion, and Emu2. In terms of layout control, MoGen strictly adheres to textual and spatial constraints, ensuring precise object quantities and accurate bounding box alignment with reasonable scene layouts. In contrast, Bounded-attention and MS-diffusion frequently exhibit positional deviations, loose spatial organization, or incorrect object quantities (e.g., omissions or extraneous additions), while Emu2, despite generating more natural images, suffers from severe deficiencies in constraint adherence. Moreover, regarding object reference fidelity and semantic integrity, MoGen effectively preserves the morphological and textural characteristics of input objects with strong photorealism. This surpasses tendency of Bounded-attention toward blurriness and the limited appearance fidelity observed in MS-diffusion and Emu2. Consequently, MoGen achieves optimal overall quality, balancing precise controllability with high visual realism in multi-object generation tasks.

\noindent\textit{\textbf{T+S~→~Image}} and \textit{\textbf{T+S+O~→~Image}.} As shown in Fig.~\ref{fig:comparison-text+str+obj}, MoGen demonstrates superior performance across multiple evaluation dimensions. In terms of quantity consistency, MoGen exhibits the most stable performance, strictly following quantity specifications. For spatial consistency, MoGen effectively preserves spatial relationships between input structure references and generated outputs, significantly outperforming competing methods. Regarding the balance between appearance consistency and expressive freedom, MoGen achieves the highest diversity and creativity, with MoGen (T+S+O) further enabling faithful reflection of input object references.
In contrast, MoEdit, our preceding work, shows certain advantages in single-category batch editing tasks but exhibits limitations in simultaneous multi-category editing and complex scene generation. MoGen systematically addresses these deficiencies, achieving effective extension from editing to free generation. Meanwhile, StableFlow frequently suffers from severe deviations in quantity consistency, producing results with missing or redundant objects. Although FlowEdit demonstrates structural preservation to some extent, it exhibits notable instability with semantic mismatches and visual inconsistencies. Overall, MoGen, particularly MoGen (T+S+O), effectively addresses performance limitations of MoEdit and achieves optimal generation quality.

%\noindent\textbf{Overall.} The visualization results demonstrate that MoGen exhibits exceptional stability and expressiveness in multi-object image generation under multi-source control signals. The method effectively maintains consistency in object quantity, spatial layout, and appearance throughout the generation process, while achieving improvements in image quality and aesthetic dimensions. Compared to the prior work MoEdit, MoGen overcomes key limitations including insufficient multi-category simultaneous editing capabilities, constrained expressive freedom due to reliance on visual anchor points, and restrictions imposed by single generation configurations. This establishes a more generalizable and efficient generation paradigm.
\subsection{Ablation Study}
\label{sec:ablation_study}
\begin{figure*}[!t]
	\centering
    \large
	\begin{overpic}[width=1.0\textwidth, trim=18 0 15 18, clip]{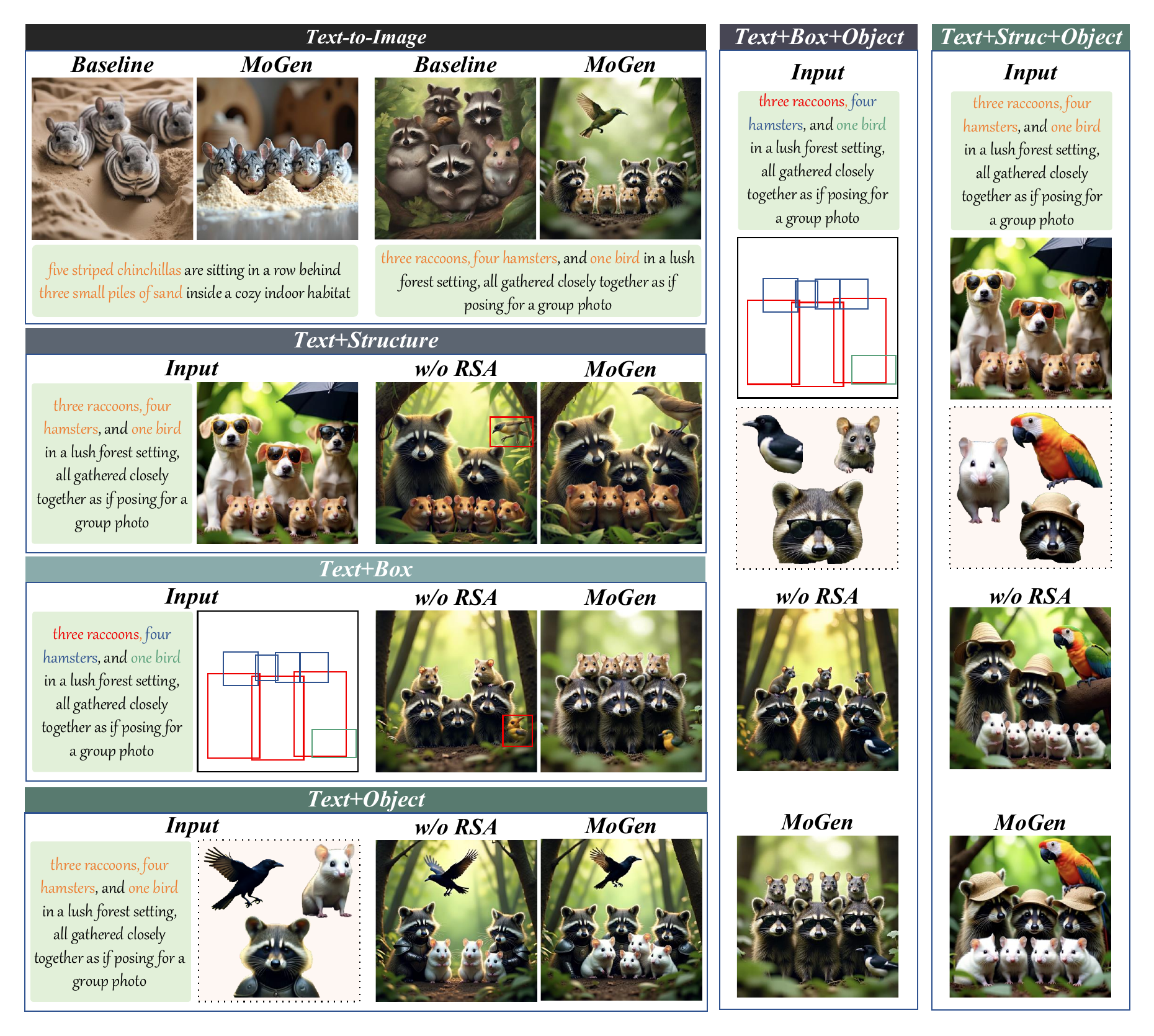}
		
	\end{overpic}
	\vspace{-25pt}
	\caption{\textbf{Ablation study}. In text-to-image, baseline SDXL exhibits quantity inconsistencies, semantic deviations, and structural confusion. In other configurations, while the AMG module provides fine-grained control, the absence of RSA (w/o RSA) leads to insufficient global semantic alignment, resulting in persistent quantity deviations. In contrast, MoGen integrates global alignment with local consistency, generating multi-object images with precise quantity, semantics, and details.}
	\label{fig:Module_effectiveness}
	\vspace{0pt}
\end{figure*}
\noindent\textbf{Effectiveness Evaluation.} As detailed in Table~\ref{tab:module_effectiveness}, the baseline model exhibits suboptimal performance across quantity consistency, semantic alignment, and image quality. Integrating the RSA module significantly mitigates semantic deviations, boosting the Numerical Accuracy from $9.31$ to $66.37$ and optimizing NIQE and DPG Score by $4.1\%$ and $8.3\%$, respectively. While the AMG module introduces essential control signals, improving NIQE by $1.8\%$ and DPG by $4.40\%$, its contribution to Numerical and Q-Align Quality is notably inferior to that of RSA due to the absence of global semantics extracted by RSA.
MoGen achieves optimal performance by synergizing both modules. Compared to the baseline, it improves NIQE by $4.6\%$, increases Numerical and DPG by $67.95$ and $9.55\%$ respectively, and secures the highest human subjective rating (MOS $+40.74$). Further comparison reveals that the MoGen yields marginal gains of $16.4\%$ (Numerical) and $12.2\%$ (MOS) over the RSA-only configuration. This demonstrates that based on RSA aligned global semantics, AMG further enhance generative consistency, collectively elevating overall generation accuracy.

\noindent\textbf{Effectiveness Demonstration.} As illustrated in Fig.~\ref{fig:Module_effectiveness}, we compare the generation performance against different module configurations:
The baseline SDXL \cite{podell2023sdxl} model, limited to $\textit{\textbf{T~→~Image}}$ generation, exhibits significant deficiencies, including inconsistent object quantities, inadequate semantic alignment, structural confusion, and poor image clarity. In other configurations, AMG module is utilized to achieve explicit constraints injection, allowing fine-grained control over scene layouts and object attributes while improving localized generation consistency. Nevertheless, without RSA (w/o RSA), the model struggles to maintain global semantic coherence, resulting in inconsistent object quantities and detail distortion. In contrast, MoGen, integrated with RSA and AMG, not only generates multi-object images with consistent quantities and refined details based on textual input, but also flexibly incorporates control signals, achieving fine-grained control and ensuring quantity consistency.
\begin{table}[!t]
    \LARGE
	\centering
	\caption{\textbf{Ablation study.} The {red} represents the {\textcolor{red}{best}}.}
	\label{tab:module_effectiveness}
	\vspace{-5pt}
	\renewcommand{\arraystretch}{1.2}
	\resizebox{\columnwidth}{!}{
		\begin{threeparttable}
			\begin{tabular}{ccccccccc}
				\hline
				\multirow{2}{*}{\textbf{RSA}} & 
				\multirow{2}{*}{\textbf{AMG}} & 
				\multirow{2}{*}{\textbf{NIQE$\downarrow$}} & 
				\multirow{2}{*}{\textbf{DPG Score$\uparrow$}} & 
				\multicolumn{2}{c}{\textbf{Q-Align}} & 
				\multirow{2}{*}{\textbf{Numerical$\uparrow$}} & 
				\multirow{2}{*}{\textbf{MOS$\uparrow$}} & 
				\multirow{2}{*}{\textbf{Signal}} \\ \cline{5-6}
				& & & & 
				\textbf{Quality$\uparrow$} & 
				\textbf{Aesthetic$\uparrow$} & & & \\ \hline
				\ding{55} & \ding{55} & 2.804 & 84.54 & 4.695 & 4.274 & 9.31 & 1.38 & No \\ 
				\ding{51} & \ding{55} & 2.688 & 91.59 & 4.793 & 4.541 & 66.37 & 37.53 & No \\ 
				\ding{55} & \ding{51} & 2.753 & 88.26 & 4.767 & 4.381 & 28.91 & 18.95 & Yes \\ 
				\lb\ding{51} & \lb\ding{51} & 
				\lb{\textcolor{red}{2.674}} & 
				\lb{\textcolor{red}{92.61}} & 
				\lb{\textcolor{red}{4.827}} & 
				\lb{\textcolor{red}{4.582}} & 
				\lb{\textcolor{red}{77.26}} & 
				\lb{\textcolor{red}{42.12}} & 
				\lb Yes \\ \hline
			\end{tabular}
		\end{threeparttable}
	}
	\vspace{-0pt}
\end{table}
\begin{figure*}[h]
	\centering
	\begin{overpic}[width=0.99\textwidth, trim=18 0 20 0, clip]{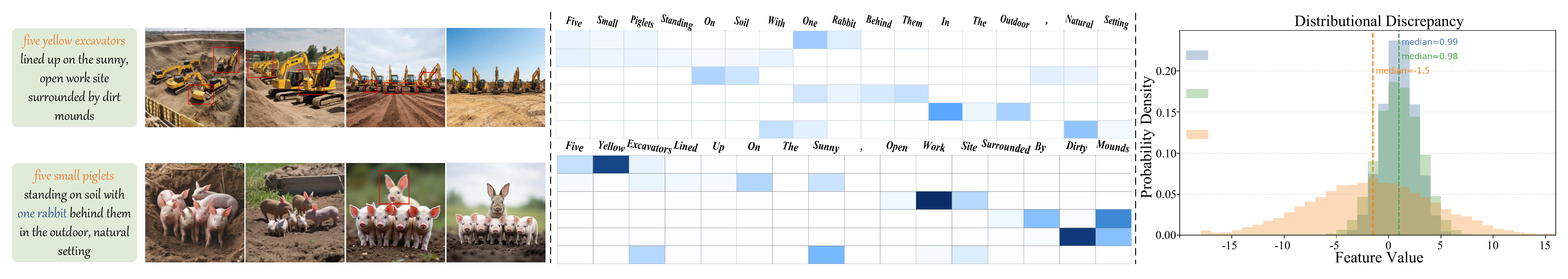}
	\put(10.1,16.7){\tiny\parbox{4cm}{Baseline}}
	\put(15.5,16.7){\tiny\parbox{4cm}{w/o $\mathbf{T}_{glob}$}}
	\put(22.5,16.7){\tiny\parbox{4cm}{w/o $\mathbf{T}_{phr}$}}
	\put(29.8,16.7){\tiny\parbox{4cm}{MoGen}}
	\put(0,-0.8){\scriptsize\parbox{6cm}{(a) Analysis of $\mathbf{T}_{glob}$ and $\mathbf{T}_{phr}$}}
	\put(34.5,-0.8){\scriptsize\parbox{9cm}{(b) Cross-attention heatmap in phrase-level branch $\mathbf{SP}_{phr}$}}
    \put(73,-0.8){\scriptsize\parbox{9cm}{(c) Feature distribution}}
    \put(77.7,14.3){\tiny\parbox{4cm}{$\mathbf{V}_{emb}$}}
    \put(77.7,11.6){\tiny\parbox{4cm}{$\mathbf{V}'_{glob}$}}
    \put(77.7,9){\tiny\parbox{4cm}{$\mathbf{V}'_{phr}$}}
	\end{overpic}
	\vspace{-0pt}
	\caption{\textbf{Analysis of the Semantic Parser}. (a) The comparison highlights that global text semantics $\mathbf{T}_{glob}$ maintain overall image structure, while phrase-level text semantics $\mathbf{T}_{phr}$ ensure localized guidance to prevent attribute aliasing and enable quantity consistency. Their synergy achieves accurate multi-object synthesis. (b) Each row represents the attention distribution of a specific query across the input text tokens. The distinct activation patterns confirm that queries effectively perform phrase chunking for precise alignment between localized generation regions and text semantics. (c) Analysis of distributional discrepancies reveals that $\mathbf{V}_{phr}'$ induces large shifts, justifying our strategy to inject it exclusively into layout blocks for enhanced stability.
}
	\label{fig:Semantic_Parser}
	\vspace{0pt}
\end{figure*}
\begin{table*}[t]
	\centering
	\scriptsize
	\caption{\textbf{Ablation study.} The {red} represents the \textcolor{red}{best}. “N/A” represents not applicable. Self-Attn(Ada.) stands for self-attention in Adaptive Controller.}
	\label{tab:ablation_analysis}
	\vspace{-5pt}
	\renewcommand{\arraystretch}{1.2}
	\resizebox{\textwidth}{!}{
		\begin{threeparttable}
			\begin{tabular}{cccccccccc}
				\hline
				% 第一行：主标题
				\multirow{2}{*}{{$\mathbf{T}_{glob}$}} & \multirow{2}{*}{$\mathbf{T}_{phr}$} & \multirow{2}{*}{\textbf{Self-Attn(Ada.)}} & \multirow{2}{*}{\textbf{DPG Score↑}} & \multicolumn{2}{c}{\textbf{Q-Align}} & \multirow{2}{*}{\textbf{IP-Sim↑}} & \multirow{2}{*}{\textbf{Spatial-Sim↑}} & \multirow{2}{*}{\textbf{Numerical↑}} & \multirow{2}{*}{\textbf{MOS↑}} \\ \cline{5-6}
				% 第二行：子标题
				& & & & \textbf{Quality↑} & \textbf{Aesthetic↑} & & & & \\ \hline

				% ===== 新增：合并整行的说明行（在第一行数据上方）=====
				\multicolumn{10}{c}{\lga\textbf{Analysis of $\mathbf{T}_{glob}$ and $\mathbf{T}_{phr}$}} \\
				\hline
				% =======================================================

				% 数据行 1
				\ding{55} & \ding{55} &  N/A &  83.91 &  4.494 &  4.172 &  N/A &  N/A &  9.03 &  4.01 \\
				% 数据行 2
				\ding{51} & \ding{55} & N/A & 87.16 & 4.591 & 4.373 & N/A & N/A & 40.38 & 18.96 \\
				% 数据行 3
				\ding{55} & \ding{51} &  N/A &  84.21 &  4.473 &  4.036 &  N/A &  N/A &  7.29 &  0.0 \\
				
				% === 新增部分开始 ===
				\lb\ding{51} & \lb\ding{51} & \lb N/A & \lb\textcolor{red}{93.01} & \lb\textcolor{red}{4.872} & \lb\textcolor{red}{4.406} &\lb N/A &\lb N/A &\lb \textcolor{red}{68.12} &\lb \textcolor{red}{77.03} \\
				\hline
                \multicolumn{10}{c}{\lga\textbf{Analysis of Self-attention in Adaptive Controller}} \\
                \hline  % 双横线
				% === 新增部分结束 ===
				
				% 数据行 4
				\ding{51} & \ding{51} & \ding{55} &  88.05 &  4.102 &  3.724 &  0.452 &  0.377 &  39.51 &  0.0 \\
				% 数据行 5
				\lb\ding{51} & \lb\ding{51} & \lb\ding{51} & \lb\textcolor{red}{92.95} & \lb\textcolor{red}{4.785} & \lb\textcolor{red}{4.514} & \lb\textcolor{red}{0.723} & \lb\textcolor{red}{0.675} & \lb\textcolor{red}{76.26} & \lb\textcolor{red}{100.0} \\ \hline
			\end{tabular}
		\end{threeparttable}
	}
	\vspace{-0pt}
\end{table*}

\noindent\textbf{Analysis of Semantic Parser.}
We first investigate the necessity and complementary nature of jointly modeling global text semantics $\mathbf{T}_{glob}$ and phrase-level text semantics $\mathbf{T}_{phr}$ within the Semantic Parser. As shown in Fig.~\ref{fig:Semantic_Parser}(a), both the baseline and w/o $\mathbf{T}_{glob}$ fail to maintain coherent global image structure in multi-object scenes, resulting in severe object fusion, quantity inconsistency, and quality degradation. Conversely, while the presence of $\mathbf{T}_{glob}$ in the w/o $\mathbf{T}_{phr}$ preserves basic image structure and quality, the absence of local guidance leads to chaotic localized image generation, manifesting as semantic omission, attribute aliasing, detail collapse, and quantity misgeneration. Only by effectively synergizing $\mathbf{T}_{glob}$ and $\mathbf{T}_{phr}$ does MoGen successfully synthesize multi-object images with accurate quantities, distinct structures, and clear attributes.
Furthermore, to validate the phrase semantics disentanglement capabilities of Semantic Parser in contexts, we visualize the cross-attention within the phrase-level attribute branch $\mathbf{SP}_{phr}$. Since the text embeddings $\mathbf{T}_{emb}$ maintain temporal alignment with input tokens, the attention scores directly reflect the focus of queries on distinct semantic units. Fig.~\ref{fig:Semantic_Parser}(b) illustrates the attention distribution of six queries across the entire sentence, where color intensity indicates the magnitude of attention weights. Observations confirm that each query effectively performs phrase chunking within the global context, establishing a solid foundation for the precise alignment between localized generation regions and specific text semantics.

Table~\ref{tab:ablation_analysis} further illustrates the interplay between $\mathbf{T}_{glob}$ and $\mathbf{T}_{phr}$. Deploying $\mathbf{T}_{glob}$ in isolation significantly enhances overall generation quality; specifically, the $3.87\%$ increase in DPG Score and the surge in the Numerical (from $9.03$ to $40.38$) validate the critical role of $\mathbf{T}_{glob}$ in maintaining global consistency and preventing structural deviation. In contrast, applying $\mathbf{T}_{phr}$ alone yields no effective gains, with Quality and Aesthetic scores declining by $0.47\%$ and $3.26\%$, respectively. The significant drop in Numerical suggests that without the global guidance of $\mathbf{T}_{glob}$, $\mathbf{T}_{phr}$ is prone to inducing instability and fragmentation. However, the joint deployment of both components generates a strong synergistic effect: all metrics outperform the other configurations, with Numerical peaking at $68.12$, thereby significantly improving the alignment with human preferences.

\noindent\textbf{Analysis of Synergistic Utilization Mechanism.}
To validate this design, we conducted an in-depth analysis of feature distributions (see Fig.~\ref{fig:Semantic_Parser}(c) and Eq.~\ref{eq:out Synergistic Utilization Mechanism}). Specifically, we compared the distributional discrepancies between features $\mathbf{V}_{glob}'$ induced by $\mathbf{T}_{glob}$ and $\mathbf{V}_{phr}'$ induced by $\mathbf{T}_{phr}$. Since $\mathbf{T}_{glob}$ inherits the pretrained cross-attention weights of the U-Net, the resulting $\mathbf{V}_{glob}'$ distribution (range $[-6, 8]$) aligns closely with that of the original distribution (range $[-4, 6]$) induced by $\mathbf{T}_{emb}$. This indicates that $\mathbf{T}_{glob}$ can be injected across all blocks with minimal perturbation to the feature space. Conversely, $\mathbf{V}_{phr}'$, derived from trainable modules, exhibits significant distribution shifts, with its range expanding to $[-18, 15]$ and the median drifting to approximately $-1.5$. Accordingly, we implemented a differential injection strategy: introducing $\mathbf{T}_{phr}$ exclusively into layout blocks (following findings from MoEdit \cite{li2025moedit}). This strategy effectively confines large-magnitude feature variations to specific blocks, enabling precise text semantics alignment while preserving pretrained performance, thereby enhancing generation stability.
% \begin{figure}[!t]
% 	\centering
% 	\begin{overpic}[width=0.50\textwidth, trim=18 0 0 0, clip]{fig/Synergistic Utilization Mechanism.pdf}
% 		\put(2,38){\small\parbox{4cm}{$t_e$ injected (original)}}
% 		\put(2,24.5){\small\parbox{4cm}{$t_g$ injected ($V_1$)}}
% 		\put(2,13.5){\small\parbox{4cm}{$t_p$ injected ($V_2$)}}
% 	\end{overpic}
% 	\vspace{-25pt}
% 	\caption{\textbf{Analysis of feature distribution discrepancies}. We compare the feature space $V_1$ induced by global semantics $t_g$ against $V_2$ induced by phrase-level attribute representations $t_p$. While $V_1$ closely aligns with the original pretrained distribution (indicating minimal perturbation), $V_2$ exhibits significant distribution shifts and range expansion. This observation motivates our differential injection strategy to balance precise controllability with generative stability.}
% 	\label{fig:Synergistic_Utilization_Mechanism}
% 	\vspace{0pt}
% \end{figure}
\begin{figure}[!t]
	\centering
	\begin{overpic}[width=0.5\textwidth, trim=18 0 0 23, clip]{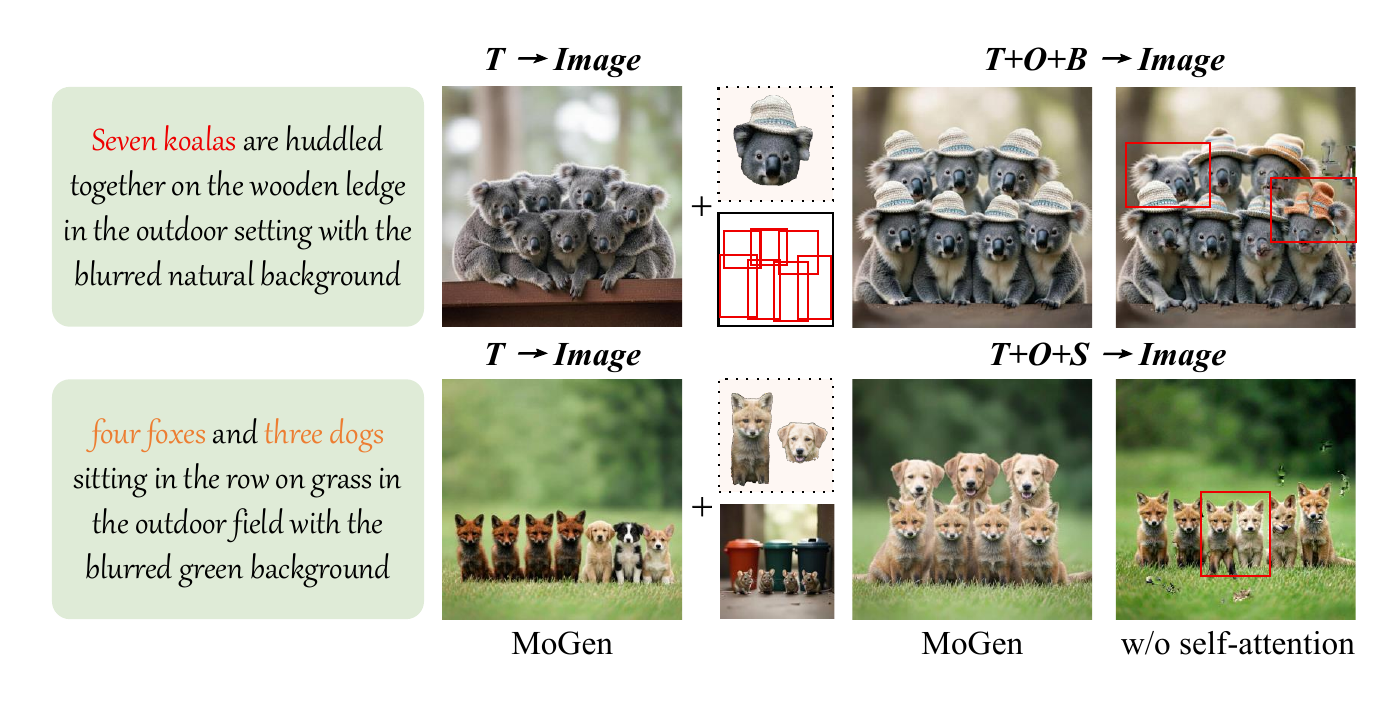}
		
	\end{overpic}
	\vspace{-25pt}
	\caption{\textbf{Ablation study on the self-attention in Adaptive Controller}. w/o self-attention leads to severe feature aliasing and attribute contamination (e.g., the missing hat in the first row and feature bleeding between the fox and dog in the second row). In contrast, MoGen leverages self-attention to establish robust dependencies, ensuring rigorous consistency in attributes, spatial layout, and object quantity.}
	\label{fig:w/o_self-attention}
	\vspace{0pt}
\end{figure}

\noindent\textbf{Analysis of Attention Mechanisms in Adaptive Controller.}
The cross-attention mechanism within Adaptive Controller is designed to dynamically extract and integrate diverse signal combinations. As illustrated in Fig.~\ref{fig:Module_effectiveness}, guided by a fixed text prompt, cross-attention maintains efficient integration capabilities when input signals vary. The subsequent self-attention mechanism precisely binds mutually disjoint constraints derived from these integrated signal semantics, significantly enhancing image quality and stabilizing consistency across attributes, spatial positioning, and quantity. 
Results from Fig.~\ref{fig:w/o_self-attention} and Table~\ref{tab:ablation_analysis} validate the efficacy of this component. Its removal precipitates severe attribute mismatch and feature entanglement, causing structural incoherence and artifacts. Quantitatively, compared to the full MoGen, excluding this mechanism degrades DPG Score and Q-Align Quality by $5.28\%$ and $6.56\%$, and Q-Align Aesthetic by $17.49\%$. Crucially, alignment metrics drop sharply: IP-Sim and Spatial-Sim decline by $37.5\%$ and $44.1\%$, respectively, causing a $48.2\%$ fall in Numerical. Visually, the first row displays attribute loss (e.g., a missing hat) and color deviation, while the second row evidences severe attribute aliasing, entangling dog and fox features, and reduced quantity consistency. Conversely, the full MoGen secures robust multi-modal dependencies, ensuring consistency across attributes, spatial layout, and object quantities.
\begin{figure}[!t]
	\centering
	\begin{overpic}[width=0.5\textwidth, trim=18 0 0 23, clip]{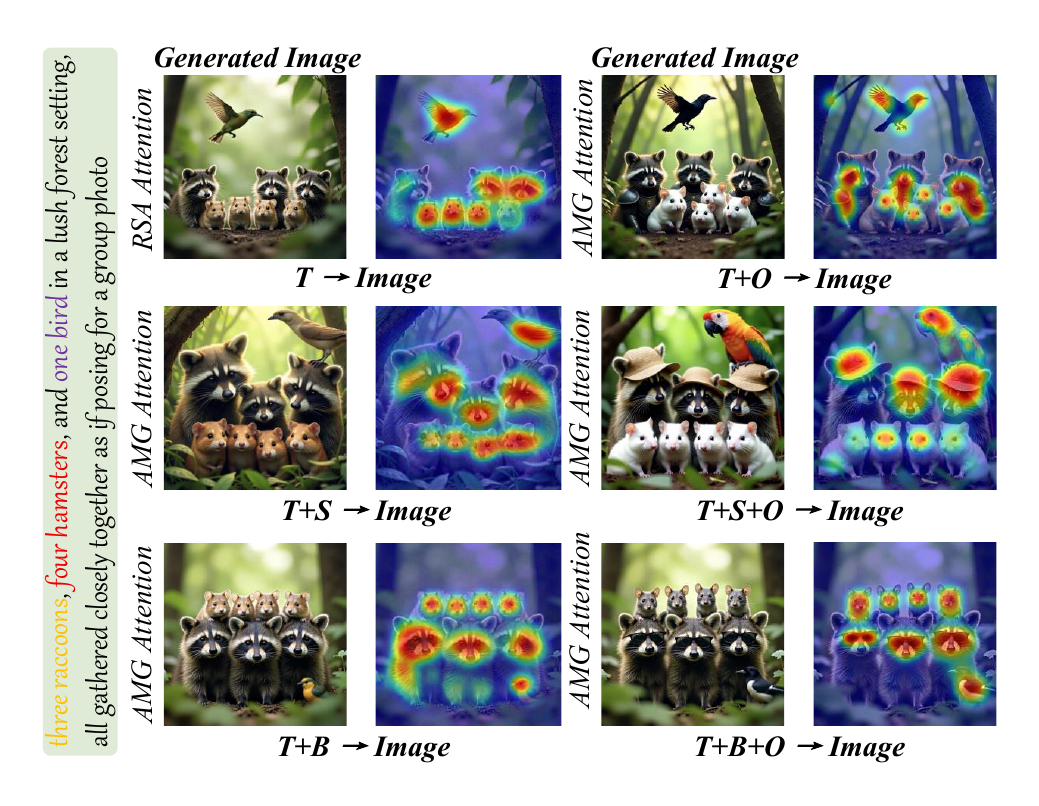}
		
	\end{overpic}
	\vspace{-20pt}
	\caption{\textbf{Interpretability analysis}. The RSA module captures globally consistent spatial relationships with strong regional coherence, ensuring quantity consistency in multi-object generation. The AMG module focuses on locally salient and texturally distinctive regions, enabling fine-grained object perception and precise control over scene layout.}
	\label{fig:attention_map}
	\vspace{-5pt}
\end{figure}

\noindent\textbf{Interpretability Analysis.} 
To elucidate the internal mechanisms, we visualize the attention distributions of the RSA and AMG modules (Fig.~\ref{fig:attention_map}). The RSA module exhibits distinct spatial selectivity and regional coherence, ensuring quantity consistency in multi-object generation by capturing necessary global spatial correlations among objects. Conversely, the AMG module focuses on spatially compact and visually salient regions, demonstrating robust local discriminability and object perception. Incorporating the input signal from Fig.~\ref{fig:Module_effectiveness}, the attention mechanism displays slight regional coherence solely in the \textit{\textbf{T+S → Image}}; in other configurations, the model consistently focuses on regions with highly discriminative visual attributes. For instance, under the \textit{\textbf{T+B+O → Image}}, attention highlights specific features such as the sunglasses of raccoons, the faces of mice, and the bodies of birds. Crucially, these salient regions constrain fine-grained appearance while precisely anchoring the spatial positions of objects. This capability enables the model to plan fine-grained objects within complex, cluttered scenes, thereby achieving precise control over both object attributes and scene layouts.

\section{Conclusion}
This paper presents MoGen, a user-friendly approach for multi-object image generation. The method introduces an RSA module that ensures precise alignment between localized generation regions and their corresponding text semantics, thereby enabling text-to-image generation that follows quantity specifications in text prompts. Building upon this foundation, an AMG module is proposed to enable flexible integration of various combinations of multi-source control signals, allowing dynamic adjustment of fine-grained control over scene layouts and object attributes. This design significantly enhances model accessibility and control flexibility, facilitating adaptation to diverse user requirements and resource constraints. Experimental results demonstrate that MoGen not only maintains quantity consistency in multi-object generation tasks but also achieves superior performance in image quality and fine-grained control compared with existing methods.

%\appendices
%\section{Proof of the First Zonklar Equation}
%Appendix one text goes here.

% you can choose not to have a title for an appendix
% if you want by leaving the argument blank
%\section{}
%Appendix two text goes here.

%\ifCLASSOPTIONcompsoc
%  \section*{Acknowledgments}
%\else
%  \section*{Acknowledgment}
%\fi
%The authors would like to thank...
%\ifCLASSOPTIONcaptionsoff
%  \newpage
%\fi

% trigger a \newpage just before the given reference
% number - used to balance the columns on the last page
% adjust value as needed - may need to be readjusted if
% the document is modified later
%\IEEEtriggeratref{8}
% The "triggered" command can be changed if desired:
%\IEEEtriggercmd{\enlargethispage{-5in}}

% references section

% can use a bibliography generated by BibTeX as a .bbl file
% BibTeX documentation can be easily obtained at:
% http://mirror.ctan.org/biblio/bibtex/contrib/doc/
% The IEEEtran BibTeX style support page is at:
% http://www.michaelshell.org/tex/ieeetran/bibtex/
%\bibliographystyle{IEEEtran}
% argument is your BibTeX string definitions and bibliography database(s)
%\bibliography{IEEEabrv,../bib/paper}
%
% <OR> manually copy in the resultant .bbl file
% set second argument of \begin to the number of references
% (used to reserve space for the reference number labels box)

{   
	\bibliographystyle{plain}
	\bibliography{main}
}
\begin{IEEEbiography}[{\includegraphics[width=1in,height=1.25in,clip,keepaspectratio]{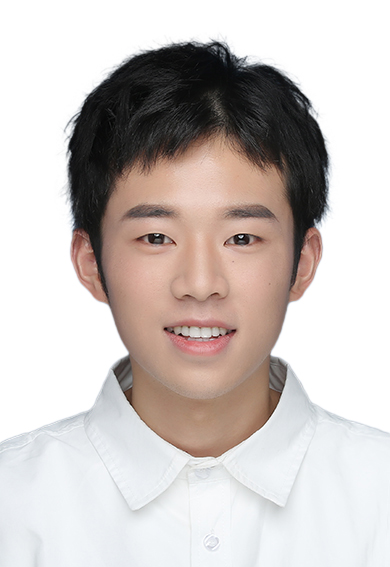}}]{Yanfeng Li} received the B.Eng. degree in Computer Science and Technology from Shanghai Polytechnic University, China, and the M.Eng. degree from the School of Information Science and Engineering, Ningbo University, China. He is currently pursuing the Ph.D. degree with the Faculty of Applied Sciences, Macao Polytechnic University, Macao, China. His current research interests include machine learning, computer vision, image processing, and the generation of images and 3D models.
\end{IEEEbiography}

\begin{IEEEbiography}[{\includegraphics[width=1in,height=1.25in,clip,keepaspectratio]{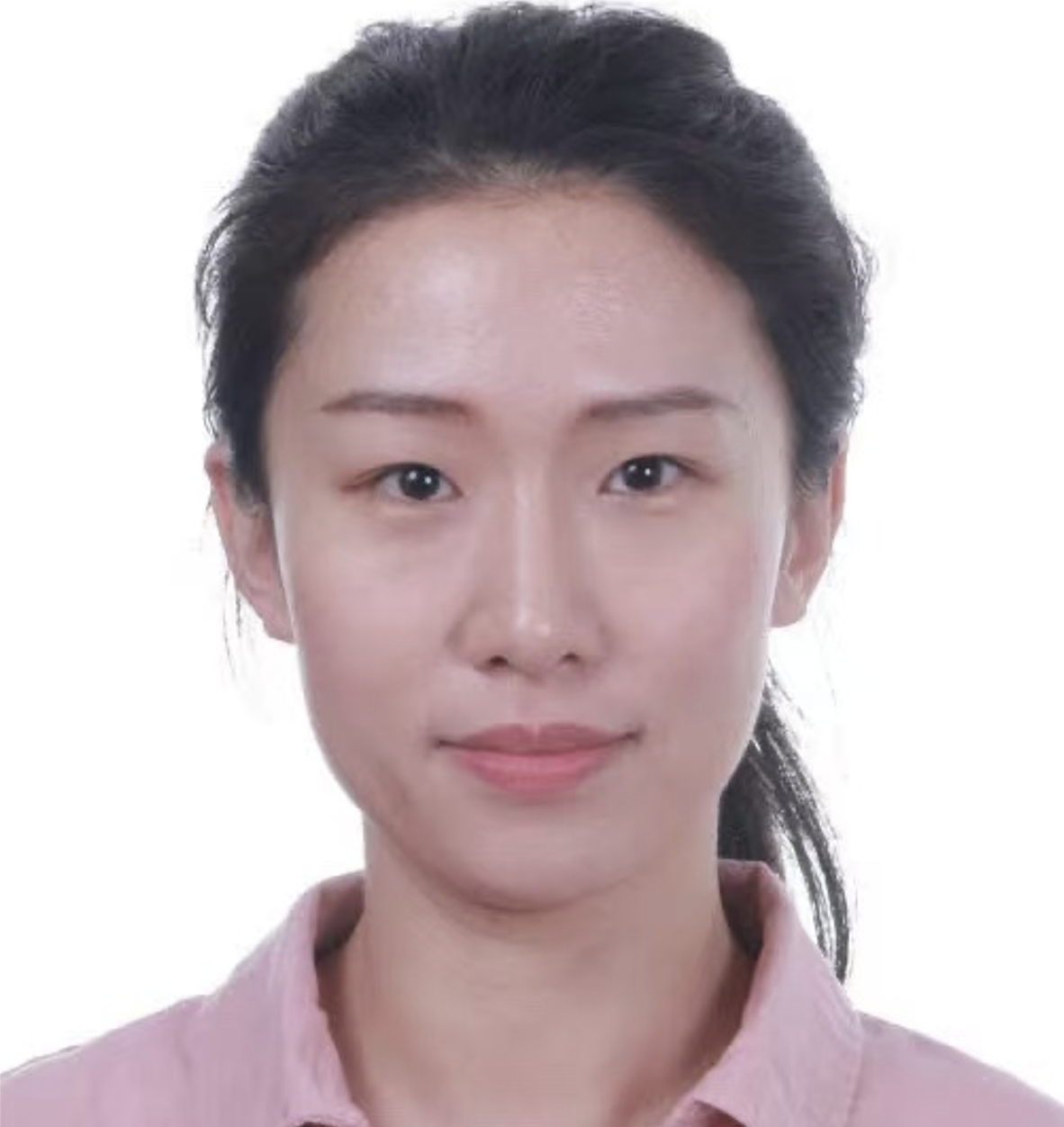}}]{Yue Sun} received the B.Sc. degree from Huazhong University of Science and Technology, the M.Sc. degree from the University of Western Ontario, and the Ph.D. degree from Eindhoven University of Technology. Her research interests include video/image processing, machine learning, pattern recognition, and related applications.
\end{IEEEbiography}

\begin{IEEEbiography}[{\includegraphics[width=1in,height=1.25in, clip,keepaspectratio]{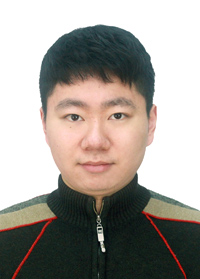}}]{Keren Fu} is currently a research associate professor with College of Computer Science, Sichuan University, Chengdu, China. He received the dual Ph.D. degrees from Shanghai Jiao Tong University, Shanghai, China, and Chalmers University of Technology, Gothenburg, Sweden. He has published over 80 papers in IEEE journals like IEEE TPAMI, TIP, TNNLS and international conferences like CVPR, ECCV, AAAI, and has won the Best Paper Honorable Mention Award at CVM Journal in 2023. His current research interests include computer vision, object detection/segmentation, and deep learning.
\end{IEEEbiography}

\begin{IEEEbiography}[{\includegraphics[width=1in,height=1.25in, clip,keepaspectratio]{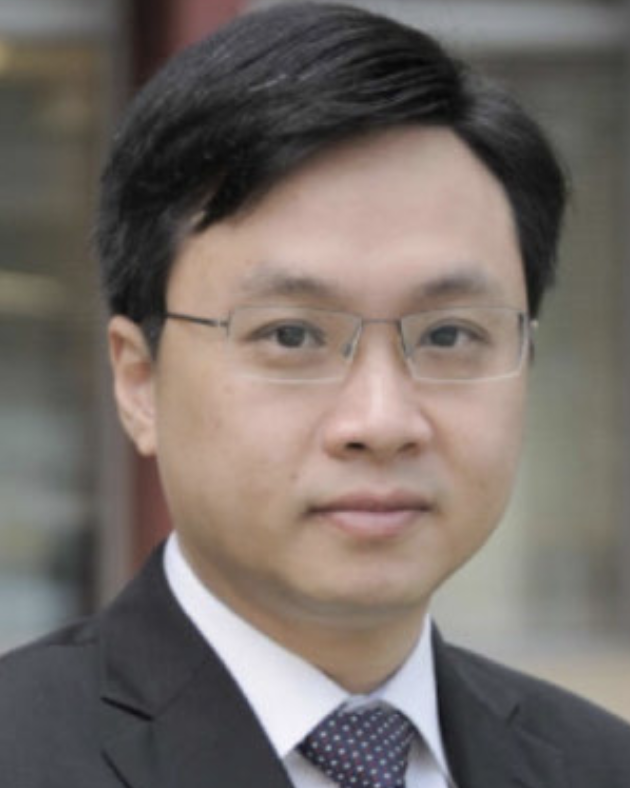}}]{Sio-Kei Im} (Senior Member, IEEE) eceived the degree in computer science and the master’s degree in enterprise information systems from the King’s College, University of London, U.K., in 1998 and 1999, respectively, and the Ph.D. degree in electronic engineering from the Queen Mary University of London (QMUL), U.K., in 2007. He gained the position of a Lecturer with the Computing Program, Macao Polytechnic Institute (MPI), in 2001. In 2005, he became the Operations Manager of MPI-QMUL Information Systems Research Center, jointly operated by MPI and QMUL, where he carried out signal processing work. He was promoted to a Professor with MPI, in 2015. He was a Visiting Scholar with the School of Engineering, University of California at Los Angeles (UCLA), and an Honorary Professor with the Open University of Hong Kong.
\end{IEEEbiography}

\begin{IEEEbiography}[{\includegraphics[width=1in,height=1.25in, clip,keepaspectratio]{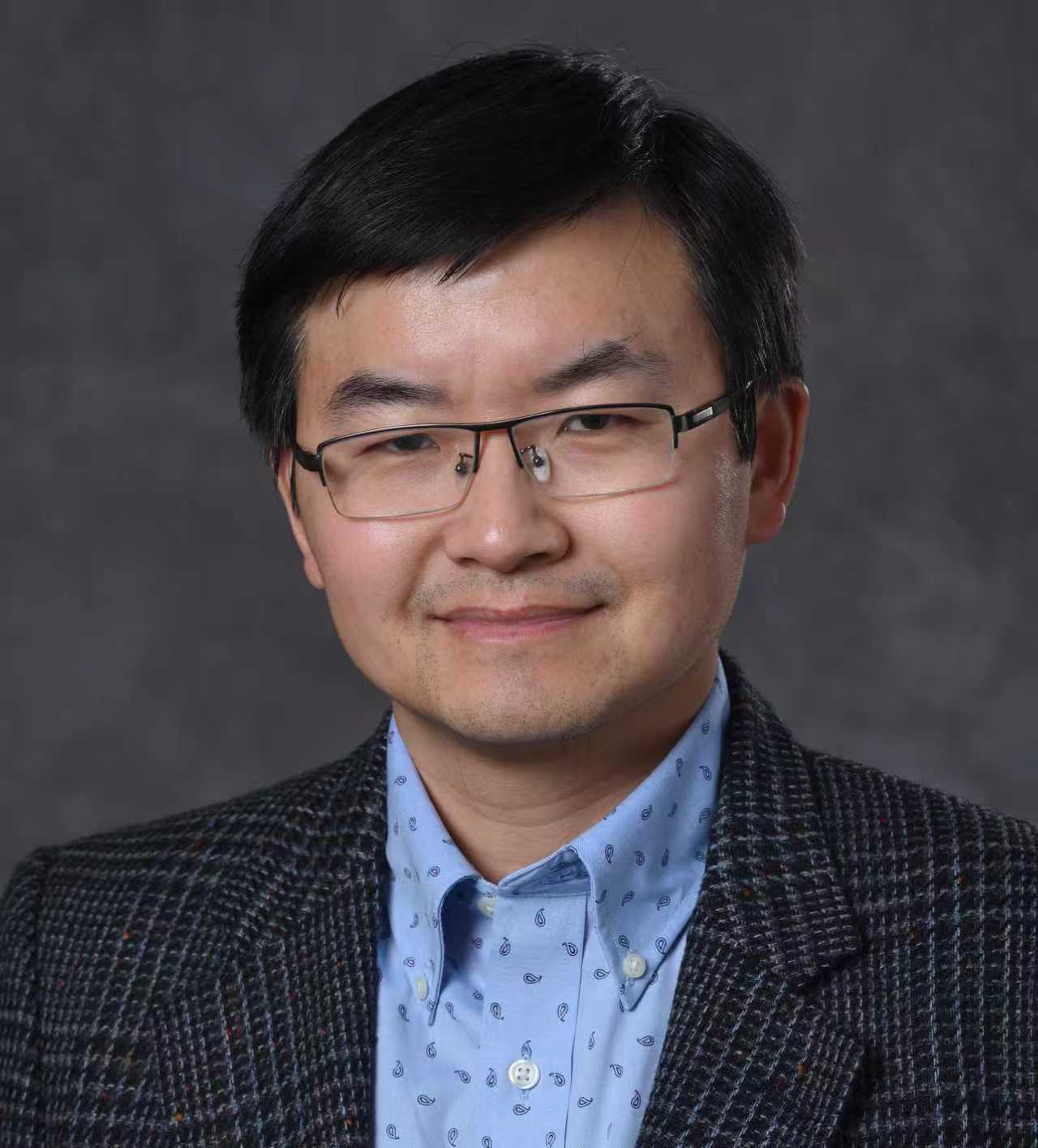}}]{Xiaoming Liu} (Fellow, IEEE) is the MSU Foundation Professor, and Anil and Nandita Jain Endowed Professor at the Department of Computer Science and Engineering of Michigan State University (MSU). He received Ph.D. degree from Carnegie Mellon University in 2004. Before joining MSU in 2012 he was a research scientist at General Electric (GE) Global Research. He works on computer vision, machine learning, and biometrics especially on face related analysis and 3D vision. Since 2012 he helps to develop a strong computer vision area in MSU who is ranked top 15 in US according to csrankings.org. He is an Associate Editor of IEEE Transactions on Pattern Analysis and Machine Intelligence. He has authored more than 200 scientific publications, and has filed 35 U.S. patents. His work has been cited over 3k times according to Google Scholar, with an H-index of 84. He is a fellow of The Institute of Electrical and Electronics Engineers (IEEE) and International Association for Pattern Recognition (IAPR).
\end{IEEEbiography}

\begin{IEEEbiography}[{\includegraphics[width=1in,height=1.25in, clip,keepaspectratio]{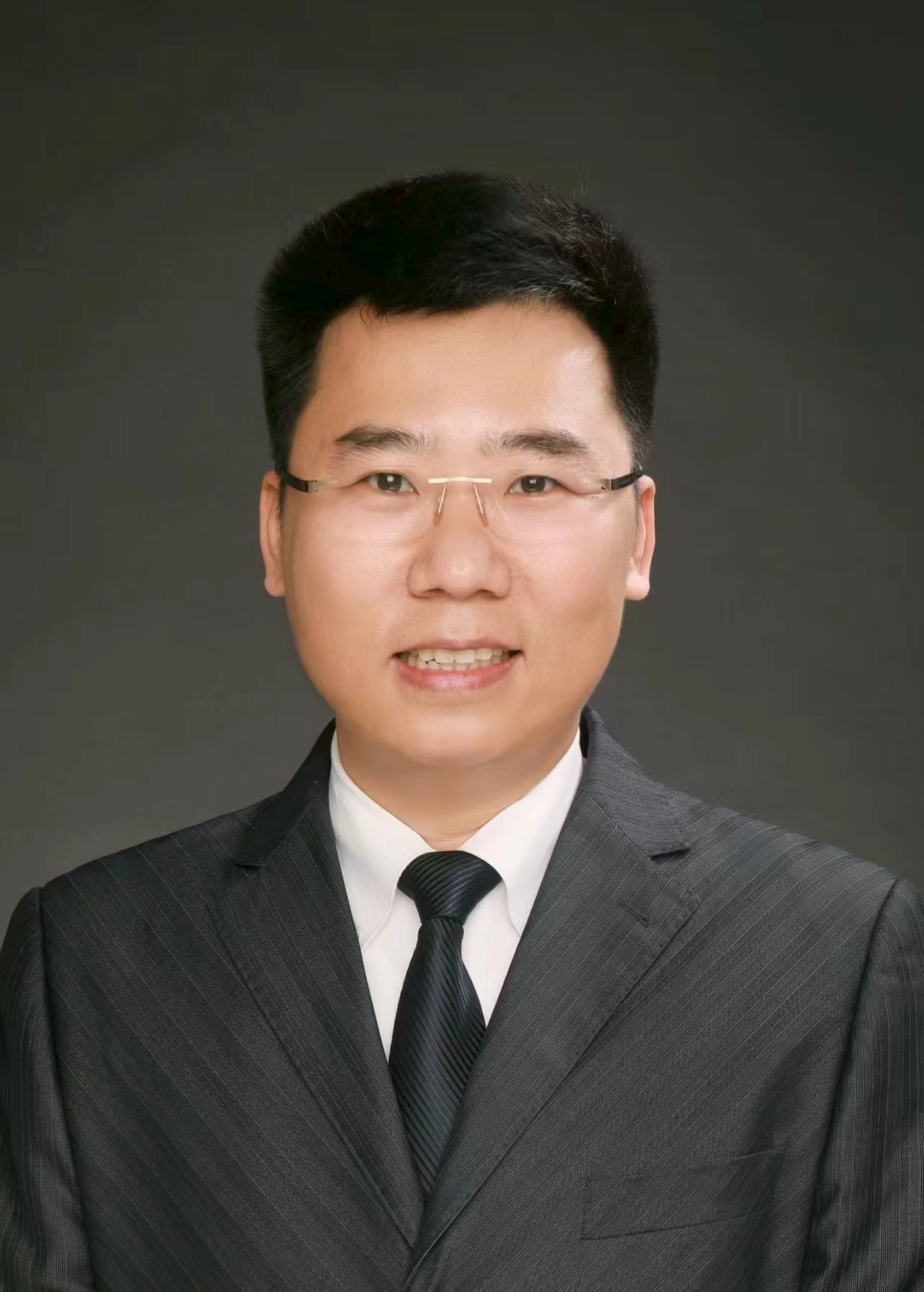}}]{Guangtao Zhai} (Fellow, IEEE) received the B.E. and M.E. degrees from Shandong University, Shandong, China, in 2001 and 2004, respectively, and the Ph.D. degree from Shanghai Jiao Tong University, Shanghai, China, in 2009. From 2008 to 2009, he was a Visiting Student with the Department of Electrical and Computer Engineering, McMaster University, Hamilton, ON, Canada, where he was a Post-Doctoral Fellow, from 2010 to 2012. From 2012 to 2013, he was a Humboldt Research Fellow with the Institute of Multimedia Communication and Signal Processing, Friedrich-Alexander-University of Erlangen–Nüremberg, Germany. He is currently a Professor with the School of Information Science and Electronic Engineering, Shanghai Jiao Tong University. His research interests include multimedia signal processing and perceptual signal processing. He received the Award of National Excellent Ph.D. Thesis from the Ministry of Education of China in 2012.
\end{IEEEbiography}

\begin{IEEEbiography}[{\includegraphics[width=1in,height=1.25in,clip,keepaspectratio]{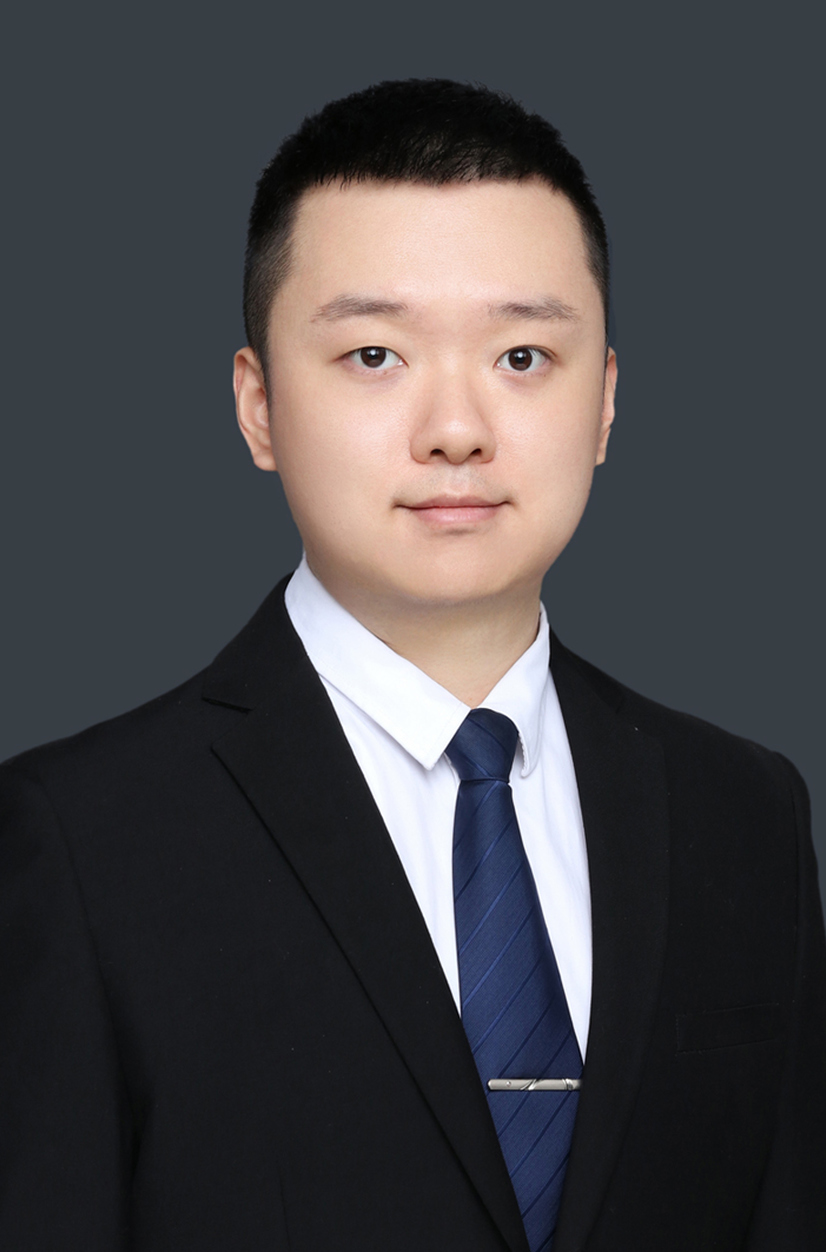}}]{Xiaohong Liu} (Member, IEEE) received the Ph.D. degree in electrical and computer engineering from McMaster University, Canada, in 2021. He is currently an associate professor with the School of Computer Science, Shanghai Jiao Tong University, China. His research interests lie in computer vision and multimedia. He has published over 100 academic papers in top journals and conferences, and has been recognized among the Stanford Top 2\% Scientists (2025), the Microsoft Research Asia StarTrack Scholars (2024), Chinese Government Award for Outstanding Self-financed Students Abroad (2021), and Borealis AI Fellowships (2020). He serves as an associate editor of the ACM TOMM.
\end{IEEEbiography}

\begin{IEEEbiography}[{\includegraphics[width=1in,height=1.25in,clip,keepaspectratio]{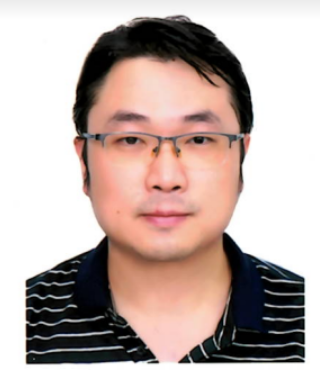}}]{Tao Tan} (Member, IEEE) received the B.Eng. degree in Biomedical Engineering from Zhejiang University, China, in 2007, the M.Sc. degree in Biomedical Engineering from Eindhoven University of Technology, the Netherlands, in 2008, and the Ph.D. degree in Radiology from Radboud University Nijmegen, the Netherlands, in 2014. He is currently an Associate Professor with the Faculty of Applied Sciences, Macao Polytechnic University, Macao, China, where he also serves as the Director of the Intelligent Medical Computing Laboratory. His research interests include medical image analysis, computer vision, and deep learning for healthcare applications.
\end{IEEEbiography}

% You can push biographies down or up by placing
% a \vfill before or after them. The appropriate
% use of \vfill depends on what kind of text is
% on the last page and whether or not the columns
% are being equalized.

%\vfill

% Can be used to pull up biographies so that the bottom of the last one
% is flush with the other column.
%\enlargethispage{-5in}

% that's all folks
\end{document}